%% file: collas2025_conference.tex
\documentclass{article} %
\usepackage{collas2025_conference,times}
\usepackage{easyReview}

\input{math_commands.tex}

\usepackage{hyperref}
\hypersetup{
    hidelinks,
    linkcolor=red,
    filecolor=magenta,
    urlcolor=blue,
    citecolor=purple,
    pdftitle={Overleaf Example},
    pdfpagemode=FullScreen,
    }

\usepackage{authblk}
\usepackage[ruled,vlined]{algorithm2e}
\usepackage{wrapfig}
\usepackage[most]{tcolorbox}
\usepackage{xcolor} %
\usepackage{booktabs}
\usepackage{graphicx}
\graphicspath{{./img/}}
\usepackage{soul}
\sethlcolor{yellow}

\newcommand{\rbar}{\bar{R}}
\newcommand{\alpharb}{\alpha_{\bar{R}}}
\newcommand{\targetent}{\bar{\mathcal{H}}}

\title{Learning Without Time-Based Embodiment Resets\\in Soft-Actor Critic}

\author[ ]{
  \textbf{Homayoon Farrahi}$^{1,2}$ \hspace{2cm}
  \textbf{A.\ Rupam Mahmood}$^{1,2,3}$
}

\affil[ ]{
  $^{1}$University of Alberta \hspace{1.5cm}
  $^{2}$Amii \hspace{1.5cm}
  $^{3}$CIFAR Canada AI Chair
}

\affil[ ]{\normalfont\texttt{\{farrahi, armahmood\}@ualberta.ca}}

\collasfinalcopy %

\begin{document}

\maketitle

\begin{abstract}

When creating new reinforcement learning tasks, practitioners often accelerate the learning process by incorporating into the task several accessory components, such as breaking the environment interaction into independent episodes and frequently resetting the environment.
Although they can enable the learning of complex intelligent behaviors, such task accessories can result in unnatural task setups and hinder long-term performance in the real world.
In this work, we explore the challenges of learning without episode terminations and robot embodiment resets using the Soft Actor-Critic (SAC) algorithm.
To learn without terminations, we present a continuing version of the SAC algorithm and show that, with simple modifications to the reward functions of existing tasks, continuing SAC can perform as well as or better than episodic SAC while reducing the sensitivity of performance to the value of the discount rate $\gamma$.
On a modified Gym Reacher task, we investigate possible explanations for the failure of continuing SAC when learning without embodiment resets.
Our results suggest that embodiment resets help with exploration of the state space in the SAC algorithm, and removing embodiment resets can lead to poor exploration of the state space and failure of or significantly slower learning.
Finally, on additional simulated tasks and a real-robot vision task, we show that increasing the entropy of the policy when performance trends worse or remains static is an effective intervention for recovering the performance lost due to not using embodiment resets.

\end{abstract}

\section{Introduction} \label{sec:introduction}

Reinforcement learning tasks often contain accessory components that are not essential to the core problem of interest, but are nonetheless integral for efficient learning by current algorithms.
Examples of such task accessories include limiting the state space with artificial boundaries, randomizing the initial states, shaping the reward functions, and choosing a suitable episode length.
Such task accessories are not essential to the definition of the core problem of interest since the same problem can be validly captured with many different choices of or, in some cases, even without the help of such accessories.
For example, the core problem of hopping forward with a one-legged robot can, in principle, be captured with or without state limitations that prevent the robot from falling, randomization of the initial states, dense compound reward functions, or variable-length episodes.
Although specific choices of task accessories can significantly accelerate the learning process, hand-designing and tuning such accessories is impractical for every new task, for example, when encountered by a continual learning agent.

Two task accessories commonly used in continuous-control tasks to accelerate learning include breaking the interaction with the environment into a sequence of episodes and resetting the embodiment of the simulated robot, which often, but not necessarily, are used in tandem.
Existing continuous-control benchmarks such as Gym Mujoco (Brockman et al.\ 2016) and Deepmind control suite (Tunyasuvunakool et al.\ 2020) most often use the episodic task setup, where each episode termination is immediately followed by an embodiment reset.
In the Gym Hopper task, for example, an episode terminates when the robot falls down, its torso tilts backward, or $1000$ steps have passed from the episode start.
Upon episode termination, the embodiment of the simulated hopper robot is instantaneously moved to a fixed standing position at the origin with small random noise added.
Although episode terminations and embodiment resets most often coincide in the episodic setup, they are independent events and can occur in isolation.
For instance, embodiment resets can occur multiple times within a single episode (Vasan et al.\ 2024), where each reset can be treated as an additional state transition that is added to the environment to accelerate learning by encouraging better exploration of the state space (Sharma et al.\ 2022).
Embodiment resets can in addition be used in the continuing task setup (Zhang \& Ross 2021) where the environment interaction happens through a single stream of experience and episode terminations do not exist by definition.

Despite accelerating the learning process, episode terminations and embodiment resets remain undesirable accessories that can result in unnatural task setups and hinder long-term performance in the real world.
In the episodic setup, interaction with the environment happens through a sequence of i.i.d.\@ episodes or trials, where the initial state of each new episode is sampled randomly and independently of the terminal state of the previous episode.
However, biological animals in nature interact with the environment through a continuous stream of experience without interruptions, in which no state is independent of past states.
In addition, unlike in simulation, executing embodiment resets in the real world takes time, during which an agent cannot influence the environment through its actions.
Thus, if the agent is evaluated continuously in real time, frequent and lengthy embodiment resets can lead to reduced asymptotic performance over time.
Finally, hand-designing and tuning the conditions and frequencies of episode terminations and embodiment resets for every new task is impractical since access to humans is not guaranteed, for instance, for an agent that learns continually.

In this work, we aim to understand the challenges of learning without episode terminations and time-based embodiment resets using the soft actor-critic (SAC, Haarnoja et al.\ 2018) algorithm.
To allow for learning without terminations, we present a continuing version of the SAC algorithm in section \ref{sec:continuing} and show that the original reward functions of some existing Gym benchmark tasks may lead to learning of degenerate behavior in the continuing setup.
Our results show that, with simple modifications to the reward functions of some tasks, the performance of continuing SAC can match or exceed that of episodic SAC and be less sensitive to the value of the discount rate $\gamma$.
On a modified Gym Reacher task, we demonstrate in section \ref{sec:statistics} that removing embodiment resets results in failure of or significantly slower learning with continuing SAC and identify poor exploration of the state space as the likely explanation of learning failure.
In section \ref{sec:interventions}, we test different interventions, including layer normalization and performance-based adjustment of the temperature parameter of SAC, for recovering the lost performance when learning without time-based embodiment resets and validate the best-performing interventions on additional simulated tasks and a real-robot vision task.

The contributions of our work are threefold.
We transform the SAC algorithm and existing tasks to the continuing setup with minimal modifications in the algorithm and the reward functions of tasks and show that the transformation can potentially improve the learning performance and reduce the sensitivity of performance to $\gamma$.
We demonstrate that a fixed or learning action-value function in the SAC algorithm can lead to poor exploration of the state space and significantly slower or failure of learning without embodiment resets.
Finally, we show that applying layer normalization to the critic network and performance-based adjustment of the temperature parameter of SAC can recover the lost performance when learning without time-based embodiment resets.
The source code is publicly available.\footnote{Code is available at \url{https://github.com/homayoonfarrahi/embodiment-resets-study}.}

\section{Problem Setup}

We use the standard reinforcement learning (RL, Sutton \& Barto 2018) setup where an agent interacts with an environment through a sequence of increasing time steps $t \geq 0$.
At each time step $t$, the agent receives the state $S_t$, performs an action $A_t$ sampled from the policy $\pi(\cdot \mid S_t)$, and receives the reward $R_{t+1}$.
The episodic setup is one where the interaction with the environment happens through a sequence of independent episodes or trials.
Each episode starts at $t=0$ and terminates at time step $T > 0$, and the goal is to maximize the expected value of the undiscounted episodic return $\int_s d_0(s)\ \mathbb{E}_\pi [\sum_{t=1}^T R_t \mid S_0=s]$, where $d_0$ denotes the starting-state distribution, from which initial states are randomly sampled.
The continuing setup is one where the interaction with the environment happens through a single uninterrupted stream of experience without terminations.
The goal of the continuing setup can be to maximize either the average-reward criterion $r(\pi) \doteq \lim_{h \to \infty} \frac{1}{h} \sum_{t=1}^h \mathbb{E} [R_t \mid A_{0:t-1} \sim \pi]$ or the discounted-reward criterion $\int_s d_\pi (s)\ v_\pi^\gamma (s) \doteq \int_s d_\pi (s)\ \mathbb{E}_\pi [\sum_{k=0}^\infty \gamma^k R_{t+k+1} \mid S_t=s]$ with $d_\pi$ as the steady-state distribution of $\pi$.
Although we report an estimate of the average-reward criterion in our results for the continuing problem setup, the solution method (SAC algorithm) that we adopt uses a discount rate $\gamma<1$ and maximizes the discounted-reward criterion.

This work aims to study the challenges of learning continuous control when only time-based embodiment resets are removed.
In addition to time-based embodiment resets, some tasks reset the embodiment of the simulated robot based on the state of the environment, such as when a one-legged robot falls down.
Learning without state-based embodiment resets, such as learning to stand up after falling down, is significantly more challenging, if not impossible, using the original reward functions of existing tasks and is out of the scope of our work.
Accordingly, we make no changes in how tasks employ state-based resets and show that learning without time-based resets may fail even when using state-based resets.
In addition, during time-based or state-based embodiment resets, we only reset the embodiment of the simulated robots and do not change the state of other entities in the environment.
Thus, we change the position of the goal in goal-reaching tasks only when the goal is reached and not during resets.

\section{Transforming SAC and Existing Tasks to the Continuing Setup} \label{sec:continuing}

In this section, we address the first challenge of transforming from the episodic to the continuing task setup.
Continuous-control tasks in simulation are conventionally modeled as episodic problems.
At the beginning of each episode, the state of the environment is randomly reset based on an initial-state distribution, which results in independent and identically-distributed (i.i.d.\@) episodes or trials.
However, biological animals in nature learn from a single uninterrupted stream of experience in which no state is independent of past states.
In addition, if resets are removed in the episodic setup, episodes will no longer be i.i.d.\@ since the initial state of an episode depends on the final state of the previous episode.
Accordingly, the continuing setup, which has no episode terminations by definition, is more naturally plausible and suitable for learning without resets.
In what follows, we outline the modifications to the SAC algorithm and existing tasks to transform them to the continuing setup.

We transform the SAC algorithm to \emph{continuing SAC} with three modifications: changing the transitions where state-based resets happen, adding a running estimate of the average reward, and replacing the TD error with the differential TD error.
Although our final goal is to learn without resets, we keep using resets in this section to allow for a fair comparison of performance between episodic and continuing setups.
In the episodic setup, an episode is terminated if and only if a state-based reset is invoked.
Since there are no terminations in the continuing setup, the experience of the agent, defined by the sequence $(S_t, A_t, R_{t+1})_{t \geq 0}$, must continue to $t \to \infty$ uninterrupted by state-based resets.
Thus, the first modification for the continuing setup is to change the terminal transition to be from the state before a state-based reset to the state after.
Assuming a state-based condition is met and the state-based reset is invoked at time step $t_f$, we ignore the termination signal and replace the terminal transition $(S_{t_f - 1}, A_{t_f - 1}, R_{T}, S_{T})$ with the new transition $(S_{t_f - 1}, A_{t_f - 1}, R_{t_f}, S_{t_f + 1})$, where $t_f + 1$ is the time step after reset.
The terminal state $S_{T}$ is ignored since the action $A_{T}$ has no effect during a reset.
We do not change or add any new transitions for time-based resets since, from the perspective of the agent, the task is continuing (Pardo et al.\ 2018), and only meeting state-based conditions, such as falling down, result in resets.

For the second modification, we maintain a running estimate of the average reward $\rbar$ as an exponential moving average of the received rewards $R$ (Naik et al.\ 2024).
The estimate is initialized to $\rbar_0=0$ and is updated at every time step according to $\rbar_{t+1} = (1-\alpharb) \rbar_t + \alpharb R_{t+1}$, with $\alpharb=0.0003$ as the step size of average reward.
The third modification changes the update rule for the Q function to minimize the semi-gradient squared differential TD error $J (\phi_i)$ defined as
\begin{align}
    &J_t (\phi_i) \doteq \frac{1}{2} \mathbb{E}_{(s,a,r,s') \sim \mathcal{D}} \left[ \left[ r - \rbar_{t+1} + \gamma \left( \min_{j \in \{1, 2\}} Q_{\bar{\phi}_{j,t}} (s',a') - \alpha_t \log \pi_{\theta_t} (a' \mid s') \right) - Q_{\phi_{i,t}} (s,a) \right]^2 \right] , \nonumber
    \label{eq:diff_r_q_loss}
\end{align}
where $\mathcal{D}$ is the replay buffer, $\phi_1$ and $\phi_2$ are the Q network weights, $\bar{\phi}_1$ and $\bar{\phi}_2$ are the target Q network weights, $\theta$ denotes the policy weights, $\alpha$ is the auto-tuned temperature parameter, and $a' \sim \pi_{\theta_t} (\cdot \mid s')$.
Continuing SAC uses two Q functions and two target Q functions as in the original SAC implementation (Haarnoja et al.\ 2018), which we refer to as \emph{episodic SAC}.
Using two Q functions was proposed by van Hasselt (2010) to avoid over-estimation bias of action values.
To stabilize training, Mnih et al. (2015) proposed using a target Q network, the weights of which are exponentially-moving averages of the Q network weights.
Fujimoto et al. (2018) used two Q networks and two target Q networks simultaneously for continuous action spaces.

\begin{figure}[t]
  \centering
  \includegraphics[width=1.0\textwidth]{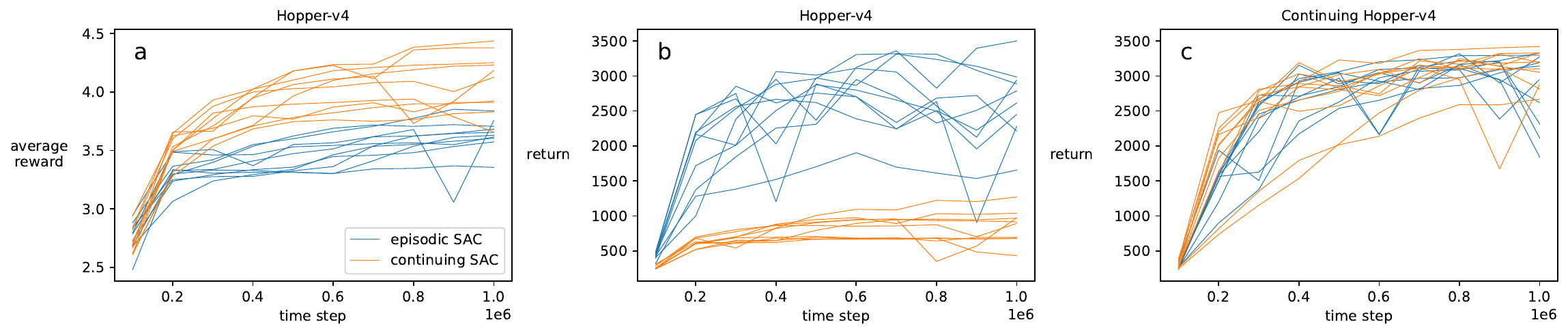}
  \caption{Although continuing SAC can achieve higher average reward than episodic SAC in most runs (a), it obtains much lower returns of less than 1000 (b). Modifying the reward function of Hopper by removing the constant term and adding a penalty for falling enables continuing SAC to perform as well as episodic SAC (c).}
  \label{fig:return_vs_rbar_term_reset_bootstrap_hopper_1x3}
\end{figure}

We ran episodic and continuing SAC on the Gym Hopper task for one million steps and stored the returns and average rewards for both algorithms.
Return was calculated after every reset as the undiscounted sum of rewards received since the previous reset.
Average reward was calculated every 1000 steps as the average of the rewards received in the past 1000 time steps.
Each line in figures \ref{fig:return_vs_rbar_term_reset_bootstrap_hopper_1x3}a and \ref{fig:return_vs_rbar_term_reset_bootstrap_hopper_1x3}b shows the average reward and return, respectively, for a single run.
Although continuing SAC can achieve higher average reward than episodic SAC in most runs in Figure \ref{fig:return_vs_rbar_term_reset_bootstrap_hopper_1x3}a, it obtains much lower returns of less than 1000 in Figure \ref{fig:return_vs_rbar_term_reset_bootstrap_hopper_1x3}b.
In Hopper, always standing still gives a return of around 1000.

The discrepancy in performance between the return and average-reward metrics is caused by a faulty environment reward function.
The original reward function of the Hopper task contains three terms: a constant reward of one except at the time step when hopper falls down, the current forward velocity, and a control cost term to penalize high action magnitudes.
Using the original reward function in the continuing setup, average reward can be maximized by repeatedly diving forward, falling, and incurring a state-based reset since diving can achieve a higher forward velocity than the intended behavior of hopping.
To fix the issue, we remove the constant reward term as it has a negligible effect on average reward and add a reward of -500 to $R_{t_f}$ as the cost of state-based resets (Zhang \& Ross 2021).
We refer to Hopper with the new reward function as \emph{continuing Hopper}.
Figure \ref{fig:return_vs_rbar_term_reset_bootstrap_hopper_1x3}c compares the returns of episodic and continuing SAC on continuing Hopper.
Although continuing Hopper is learned using its modified reward function, the returns in Figure \ref{fig:return_vs_rbar_term_reset_bootstrap_hopper_1x3}c are calculated using the original reward function of Hopper for a fair comparison of performance across environments.
As shown in Figure \ref{fig:return_vs_rbar_term_reset_bootstrap_hopper_1x3}c, the reward function modifications of continuing Hopper enable continuing SAC to perform as well as episodic SAC.

\begin{figure}[t]
  \centering
  \includegraphics[width=1.0\textwidth]{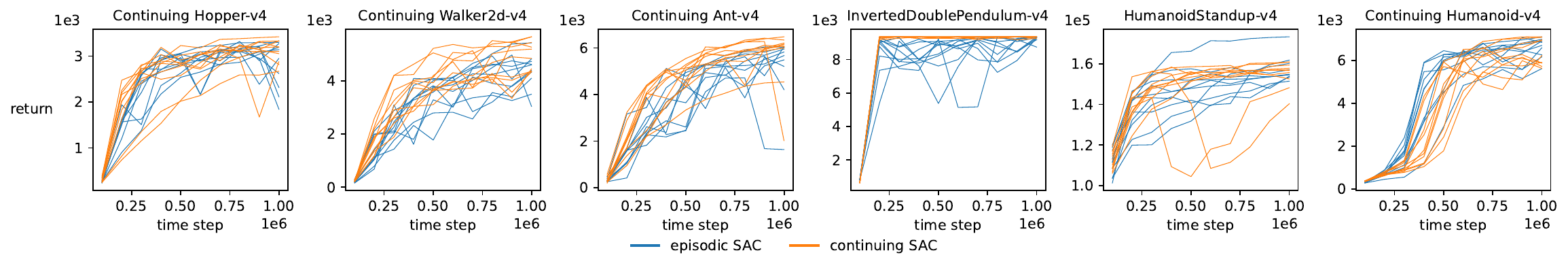}
  \caption{Continuing SAC performs similarly to episodic SAC in all tested environments. All environments use resets, which mark the end of an episode in episodic SAC and a state transition to the state after reset in continuing SAC.}
  \label{fig:return_cntg_0}
\end{figure}

\begin{figure}[ht]
    \centering
    \begin{minipage}{\linewidth}
        \centering
        \includegraphics[width=1\linewidth]{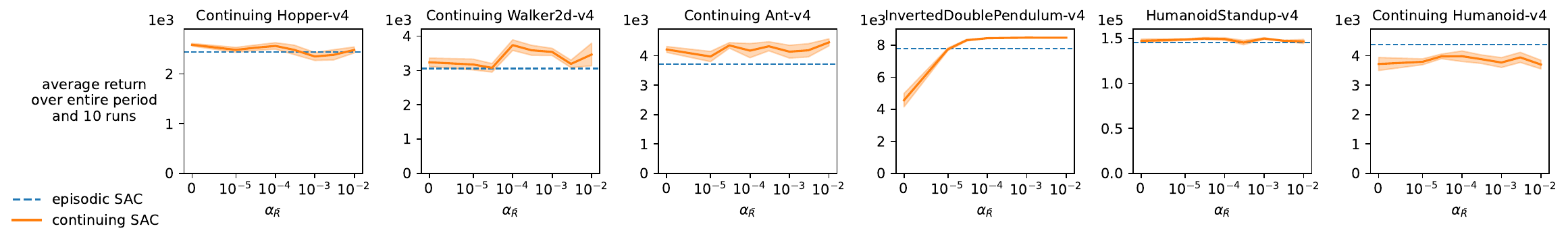}
    \end{minipage}
    \begin{minipage}{\linewidth}
        \centering
        \includegraphics[width=1\linewidth]{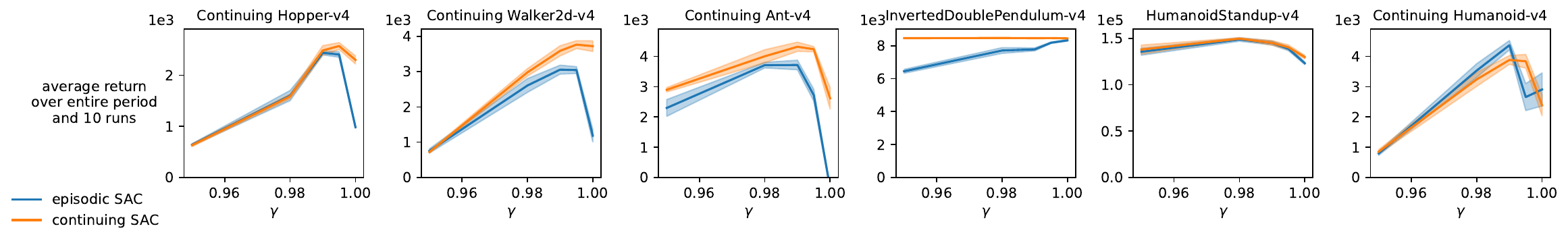}
        \caption{The performance of continuing SAC is not sensitive to the value of its hyper-parameter $\alpharb$ (top) and is less sensitive to the value of $\gamma$ than episodic SAC, particularly for $\gamma$ values close to one (bottom).}
        \label{fig:sensit_alpha_gamma}
    \end{minipage}
\end{figure}

We validated the performance of continuing SAC on additional Gym Mujoco environments with results shown in Figure \ref{fig:return_cntg_0} (see Figure \ref{fig:return_vs_rbar_term_reset_bootstrap_dm_control} for DMControl).
The reward functions of continuing Walker2d, Ant, and Humanoid were modified from their respective original reward functions by removing the constant reward term and adding state-based reset penalties of 100, 50, and 50, respectively, as was done for continuing Hopper, although, the returns in the figure were calculated using the original reward functions for ease of comparison.
As shown, continuing SAC performs similarly to episodic SAC in all tested Gym and DMControl tasks.
Furthermore, the performance of continuing SAC is not sensitive to the value of its hyper-parameter $\alpharb$ as seen in Figure \ref{fig:sensit_alpha_gamma}(top), where each point is the return averaged over the entire learning period (see Figure \ref{fig:sensit_term_reset_bootstrap_dm_control} for DMControl).
Figure \ref{fig:sensit_alpha_gamma}(bottom) compares the sensitivity of performance of episodic and continuing SAC with $\alpharb=0.0003$ to the value of the discount rate $\gamma$.
Performance of continuing SAC is less sensitive to the value of $\gamma$ than episodic SAC, particularly for $\gamma$ values close to one.

\section{The Challenge of Learning Without Time-Based Embodiment Resets} \label{sec:statistics}

In this section, we adopt the continuing task setup of section \ref{sec:continuing} and investigate the second challenge of learning without time-based embodiment resets on a modified Gym Reacher environment on which, despite the simplicity of the task, continuing SAC fails to learn without resets in many runs.
In the original Reacher, the task is to move the end-effector of a two-DOF robot arm as fast as possible to a goal position, and the position of the arm and the goal are both reset based on time every 50 time steps.
To make the Reacher task suitable for learning without resets, we apply four modifications to the task and refer to the resulting task as \emph{continuing Reacher}.
The first modification is to not change the position of the goal during resets.
The second modification is to change the position of the goal only when the distance from the fingertip to the goal and the magnitude of the velocity of the fingertip are less than a small threshold, which denotes a successful reach.
After the second modification, using a per-step reward of negative distance between the fingertip and the goal is no longer suitable in our continuing setup since successful reaches lead to a new goal position, larger distance to the goal, and lower reward.
Thus, an agent can maximize the average reward by staying close to a single goal without a successful reach.
The third modification is to use a new sparse reward function that gives a reward of 100 once for every successful reach.
The fourth modification limits the range of motion of the base joint of the robot between -3 and 3 radians to mimic the joint range limitations when learning on real robots.
The continuing Reacher task can be learned either with time-based resets every 50 time steps or without, and, with time-based resets, only the position and velocity of the arm is reset to zero with small uniform noise added.

\begin{figure}[t]
  \centering
  \includegraphics[width=0.9\textwidth]{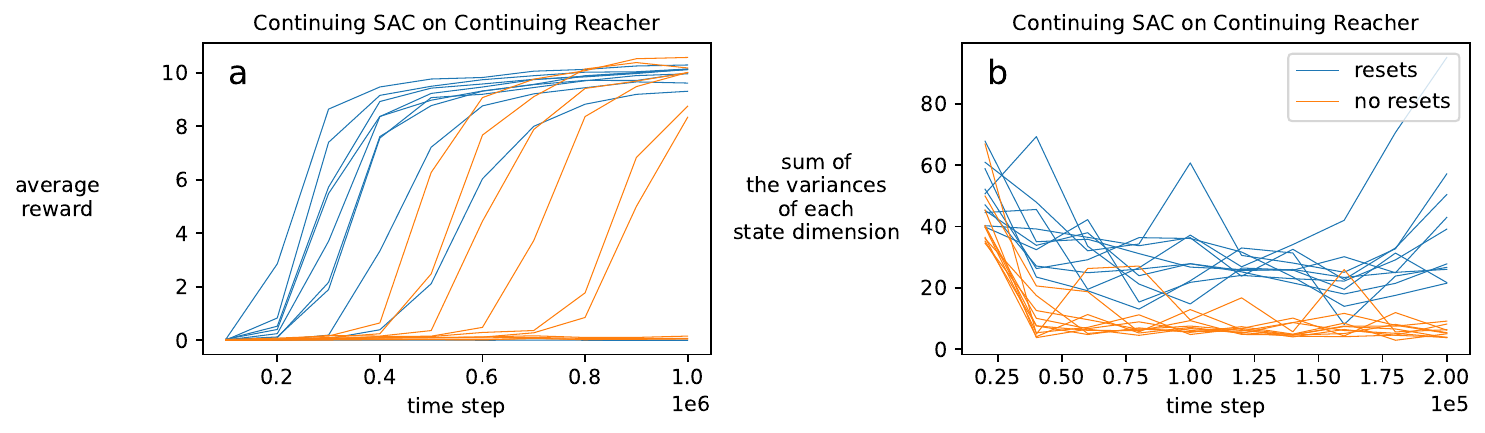}
  \caption{Not using time-based resets results in failing to learn or significantly slower learning in most runs (a) and significantly lower variance in the visited states and poor exploration of the state space (b).}
  \label{fig:r_bar_variance_ReacherNoReset-v5}
\end{figure}

We examined the impact of using time-based resets on performance by running continuing SAC on continuing Reacher with and without time-based resets for one million time steps.
Figure \ref{fig:r_bar_variance_ReacherNoReset-v5}(a) shows the average reward against time step for 10 independent runs.
As shown, not using time-based resets results in failure of or significantly slower learning in most runs.
Resetting the arm position every 50 steps might help learning by increasing the coverage of the state space, making goal reaches more likely.
In Figure \ref{fig:r_bar_variance_ReacherNoReset-v5}(b), we plot the variance of visited states in the first 200,000 time steps for the same runs from Figure \ref{fig:r_bar_variance_ReacherNoReset-v5}(a).
The variance was calculated in non-overlapping periods of 1000 steps for each state dimension separately and summed over all dimensions.
As seen in Figure \ref{fig:r_bar_variance_ReacherNoReset-v5}(b), not using time-based resets leads to significantly lower variance in the visited states and poor exploration of the state space.
When learning without resets, the robot frequently remains stationary in a fixed position or gets stuck at the limits of the base joint for long periods.

\begin{figure}[t]
  \centering
  \includegraphics[width=1.0\textwidth]{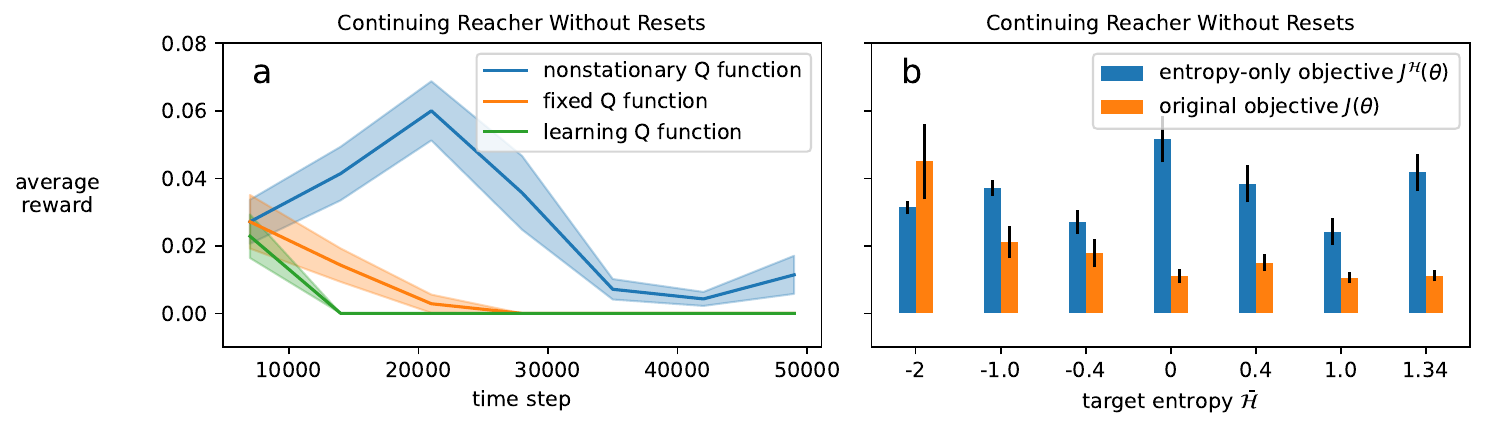}
  \caption{a) All three curves use continuing SAC without the entropy term in the policy objective. In contrast to a nonstationary Q function, fixed and learning Q functions lead to zero average reward for most of the run, indicating poor exploration, which suggests that adequate exploration of the state space in the SAC algorithm relies on a changing Q function. b) With all values except the default $\targetent=-2$, performance is higher using the entropy-only objective $J^{\mathcal{H}} (\theta)$ compared to the baseline original objective $J (\theta)$, which indicates that the entropy term alone of the policy objective of SAC is not responsible for the poor exploration when learning without resets.}
  \label{fig:converged_q_log_pi_effect_r_bar_ReacherNoReset-v5}
\end{figure}

\textbf{Are the action-value and entropy terms in the policy objective of SAC both responsible for poor exploration when learning without resets?}
At every step $t$, the SAC algorithm maximizes the policy objective
\begin{equation}
    J_t (\theta) \doteq \mathbb{E}_{s_i \sim \mathcal{D}, a_i \sim \pi_{\theta_t} (\cdot \mid s_i)} \left[ \min_{j \in \{1,2\}} Q_{\phi_{j,t}} (s_i, a_i) - \alpha_t \log \pi_{\theta_t} (a_i \mid s_i) \right] . \nonumber
\end{equation}
To understand the effect of the action-value term $\min_{j \in \{1,2\}} Q_{\phi_{j,t}} (s_i, a_i)$ independently of the entropy term $-\alpha_t \log \pi_{\theta_t} (a_i \mid s_i)$ on exploration, we ran two experiments on continuing Reacher using continuing SAC without the entropy term in the policy objective for 50,000 steps.
In the first experiment, a nonstationary Q function was randomly re-initialized at every step.
In the second experiment, a fixed Q function was randomly initialized at the beginning and fixed thereafter.
The results in Figure \ref{fig:converged_q_log_pi_effect_r_bar_ReacherNoReset-v5}(a) show that, in contrast to a nonstationary Q function, a fixed Q function leads to zero average reward for the second half of the run, indicating no goal reaches and poor exploration, similarly to the baseline of learning the Q function shown in green.
The result suggests that adequate exploration of the state space in the SAC algorithm relies on a changing Q function, and a Q function that has converged to the data in the replay buffer could lead to poor exploration of the state space.

\begin{figure}[t]
  \centering
  \includegraphics[width=1.0\textwidth]{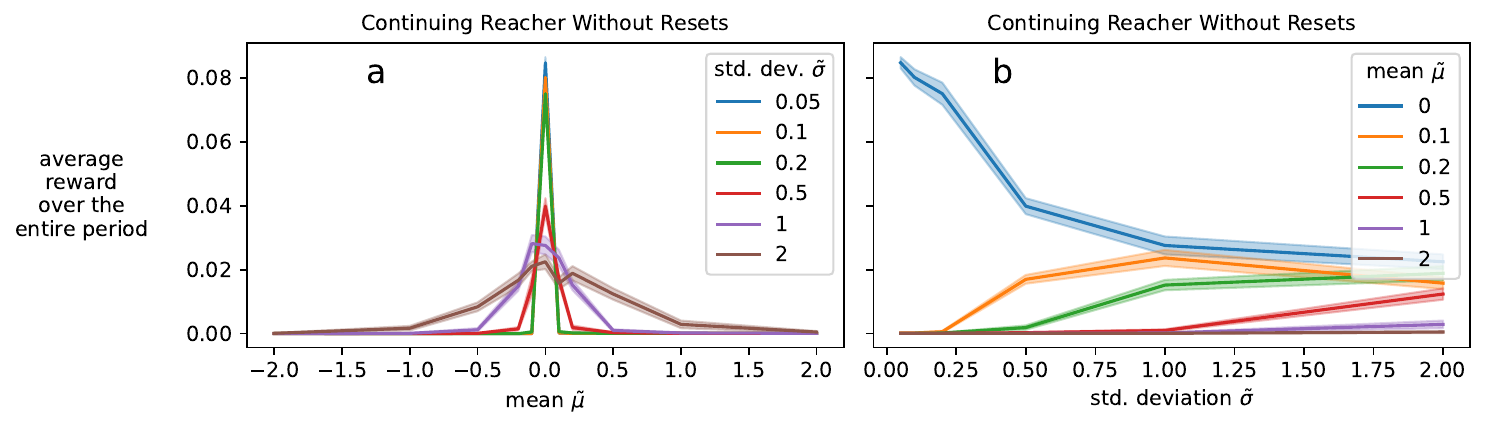}
  \caption{All curves use continuing SAC with fixed values of mean $\tilde{\mu}$ and standard deviation $\tilde{\sigma}$. a) Values of $\tilde{\mu}$ near zero yield the highest average rewards, and as $\tilde{\mu}$ moves away from zero, larger $\tilde{\sigma}$ values are needed to maintain the same average reward. b) Increasing $\tilde{\sigma}$ values generally lead to higher average rewards except for $\tilde{\mu}=0$ where smaller $\tilde{\sigma}$ values are better, likely since the fingertip needs to have low velocity for a successful reach.}
  \label{fig:no_reset_fixed_mu_sigma_r_bar_ReacherNoReset-v5}
\end{figure}

To understand the independent contribution of the entropy term to poor exploration, we removed the action-value term and used $J_t^\mathcal{H} (\theta) \doteq \mathbb{E}_{s_i \sim \mathcal{D}, a_i \sim \pi_{\theta_t} (\cdot \mid s_i)} [-\alpha_t^s \log \pi_{\theta_t} (a_i \mid s_i)]$ as the policy objective with $\alpha_t^s \doteq \alpha_t \sign \left( \mathbb{E}_{s_i \sim \mathcal{D}, a_i \sim \pi_{\theta_t} (\cdot \mid s_i)} \left[ \log \pi_{\theta_t} (a_i \mid s_i) + \targetent \right] \right)$.
We replaced the positive-valued $\alpha$ with $\alpha^s$ to ensure that the values of $\log \pi$ can get close to the target entropy $\targetent$.
Since the SAC algorithm uses squashed-gaussian action distribution, the entropy of the policy is a function of both the mean and the standard deviation.
We hypothesize that optimizing $J^\mathcal{H} (\theta)$ may thus move the mean away from zero, resulting in poor exploration.
We ran continuing SAC with $J^\mathcal{H} (\theta)$ using different values of $\targetent$ from the default value of -2 to the empirically observed maximum of 1.34 in continuing Reacher.
Figure \ref{fig:converged_q_log_pi_effect_r_bar_ReacherNoReset-v5}(b) shows the reward averaged over the entire period and 10 independent runs for different values of $\targetent$.
As shown, with all values of $\targetent$, the agent continues to reach the goal and get non-zero average reward.
Moreover, with all values except the default $\targetent=-2$, performance is higher using the entropy-only objective $J^{\mathcal{H}} (\theta)$ compared to the baseline original objective $J (\theta)$, which indicates that the entropy term alone of the policy objective of SAC is not responsible for the poor exploration when learning without resets.

\textbf{How do the values of mean $\mu$ and standard deviation $\sigma$ in the policy affect exploration in continuing Reacher?}
For each action dimension, the policy network in continuing SAC outputs $\mu_t$ and $\sigma_t$, which are then used to sample $x_t \sim \mathcal{N}(\mu_t, \sigma_t)$ and calculate an action $a_t = \tanh(x_t)$.
To understand how the output of the policy affects exploration, we ran continuing SAC on continuing Reacher for 100,000 steps, but used fixed values of mean $\tilde{\mu}$ and standard deviation $\tilde{\sigma}$ throughout the run instead of the policy outputs $\mu_t$ and $\sigma_t$.
We repeated the experiment for different fixed values of $\tilde{\mu}$ and $\tilde{\sigma}$.
Figures \ref{fig:no_reset_fixed_mu_sigma_r_bar_ReacherNoReset-v5}(a) and \ref{fig:no_reset_fixed_mu_sigma_r_bar_ReacherNoReset-v5}(b) show the reward averaged over the entire period plotted against $\tilde{\mu}$ and $\tilde{\sigma}$, respectively.
Shaded areas represent standard error.
Figure \ref{fig:no_reset_fixed_mu_sigma_r_bar_ReacherNoReset-v5}(a) shows that $\tilde{\mu}$ values close to zero yield the highest average rewards.
With large $\tilde{\mu}$ and small $\tilde{\sigma}$ values, the robot arm can get stuck at the limits of the base joint.
Figure \ref{fig:no_reset_fixed_mu_sigma_r_bar_ReacherNoReset-v5}(b) shows that increasing $\tilde{\sigma}$ values generally lead to higher average rewards except for $\tilde{\mu}=0$ where smaller $\tilde{\sigma}$ values are better, likely since the fingertip needs to have low velocity for a successful reach in continuing Reacher.

\begin{figure}[t]
    \centering
    \begin{minipage}[t]{.48\linewidth}
        \centering
        \includegraphics[width=1\linewidth]{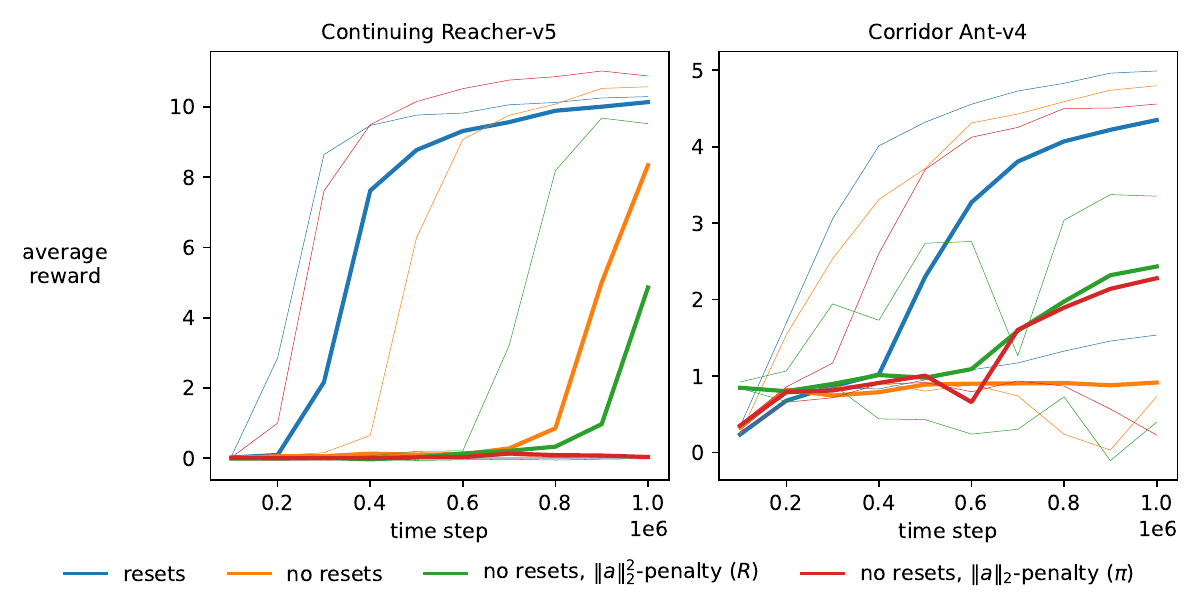}
        \caption{Neither intervention is able to match the performance of learning with resets, which suggests that large action magnitudes or saturated actions are likely not the main cause of poor exploration when learning without resets. Thick lines show the median run, and thin lines show the runs with minimum and maximum average performance over 10 independent runs.}
        \label{fig:penalties}
    \end{minipage}
    \hfill
    \begin{minipage}[t]{.48\linewidth}
        \centering
        \includegraphics[width=1\linewidth]{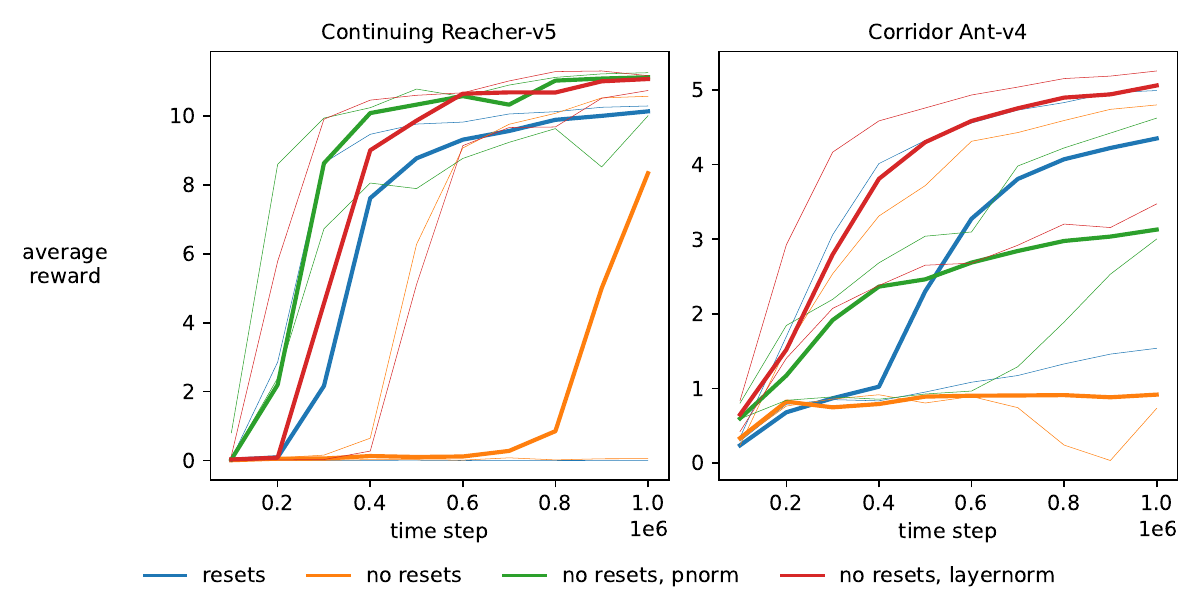}
        \caption{Both normalization techniques improve the median performance when learning without resets. When learning without resets, using layer norm can recover the performance lost due to not using resets.}
        \label{fig:norms_critic}
    \end{minipage}
\end{figure}

\section{Evaluating Different Interventions for Learning Without Time-Based Embodiment Resets} \label{sec:interventions}

In this section, we study the effect of using different interventions on recovering the performance lost when learning without resets.
We perform our experiments using continuing SAC on the continuing Reacher and a new \emph{corridor Ant} task.
The corridor Ant task was derived from the continuing Ant task in section \ref{sec:continuing} by adding two infinitely-long walls spanning along the forward direction on both sides of the initial position of Ant.
The walls put the simulated Ant robot inside a narrow corridor and mimic the spatial boundary limitations when learning on real robots.
The task can be learned with time-based resets every 1000 steps or without.
Without time-based resets, the simulated Ant robot frequently becomes stationary against the walls and does not move its joints for long periods at a time.
When the simulated ant robot rolls on its back, we use state-based resets to move ant to the initial position and orientation.
Using continuing Reacher and corridor Ant, we next explore whether penalizing large action magnitudes helps exploration.

\textbf{Does penalizing large action magnitudes improve exploration when learning without time-based resets?}
Figure \ref{fig:no_reset_fixed_mu_sigma_r_bar_ReacherNoReset-v5}a showed that choosing actions using a fixed $\tilde{\mu}=0$ leads to more goal reaches and higher average reward than $\tilde{\mu}$ values further from zero.
Values of $\tilde{\mu}$ close to zero lead to samples $x_t \sim \mathcal{N}(\tilde{\mu}, \tilde{\sigma})$ of smaller magnitudes and lower probability of saturated actions $a_t \sim \tanh(x_t)$ near $-1$ and $1$, which could help prevent the simulated robots from getting stuck at the joint limits and space boundaries.
We test whether penalizing large action magnitudes improves exploration using two interventions.
In the first intervention \emph{$\|a\|_2^2$-penalty (R)}, the value of $-\tau \|a_t\|_2^2$ is added to the reward $r_t$ at every time step with $\tau$ tuned over six values from $10^{-5}$ to $1$.
In the second intervention \emph{$\|a\|_2$-penalty ($\pi$)}, the value of $-\lambda \|a_i\|_2$ is added to the policy objective $J (\theta)$ with $\lambda$ tuned over six values from $10^{-6}$ to $0.1$ (Bjorck et al.\ 2022).
Figure \ref{fig:penalties} shows the effect of adding either of the two interventions on average reward when learning without resets.
For each experiment, we performed 10 independent runs.
After sorting the runs in ascending order of average performance over the entire period, we show the first and last runs with thin lines, and the 5th run (median) with a thick line.
Neither intervention is able to match the performance of learning with resets with only a slight improvement in median performance observed in corridor Ant.
Note that, even though corridor Ant employs state-based resets when the ant flips upside down, learning without time-based resets can still fail since the ant frequently becomes stationary against the walls for long periods at a time.
The result suggests that large action magnitudes or saturated actions are likely not the main cause of poor exploration when learning without resets.
Next, we study whether high variance in the outputs of the critic network across states contributes to poor exploration.

\textbf{Can poor exploration when learning without resets be due to high variance in outputs of the critic network across states?}
We found that saturating actions or large action magnitudes are likely not the main cause of poor exploration.
From figure \ref{fig:converged_q_log_pi_effect_r_bar_ReacherNoReset-v5}b, we also found that the entropy term alone of the policy objective of SAC is not responsible for poor exploration.
What remains is the action-value term $\min_{j \in \{1,2\}} Q_{\phi_{j,t}} (s_i, a_i)$ of the policy objective, maximizing which, we posit, results in a policy with poor exploration.
We hypothesize that the outputs of the Q function have high variance between different states-action pairs, which means some states-actions pairs have significantly higher values compared to others.
A policy that maximizes such a Q function can learn to overcommit to executing high-value actions for long periods of time.
Executing high-value actions may lead to poor exploration of the state space if all actions point toward specific states where a reward of 100 was received as in continuing Reacher, or toward states beyond the walls as in corridor Ant.
The hypothesis is consistent with the observed behavior of simulated robots remaining stationary in certain positions for long periods at a time.
We test two normalization techniques for reducing the variance of the outputs of the Q function.
The first is \emph{pnorm} (Bjorck et al.\ 2022), which divides the activation outputs of the last hidden layer by the $l2$-norm of the outputs calculated for each sample separately.
The second is \emph{layer norm} (Ba et al.\ 2016), which normalizes the pre-activation output of each hidden layer by subtracting the mean and dividing by the standard deviation of the output, where mean and standard deviation are calculated separately for each layer and sample.
Figure \ref{fig:norms_critic} shows that both normalization techniques improve the median performance when learning without resets.
On both tasks, using layer norm can recover the performance lost due to not using resets.
On corridor Ant, using layer norm results in higher median performance than pnorm.
Applying normalization to actor and critic networks simultaneously (Figure \ref{fig:norms}) yields comparable results to Figure \ref{fig:norms_critic}.
Next, we study an alternative approach for improving exploration based on the entropy term of $J (\theta)$.

\begin{wrapfigure}{r}{0.5\textwidth}
\begin{minipage}{0.5\textwidth}
\vspace{-1ex}
\begin{algorithm}[H]
\DontPrintSemicolon
\KwIn{transition dynamics $p$, initial-state distribution $d_0$, weights $\theta$, $\phi_1$, $\phi_2$, step sizes $\lambda_\theta$, $\lambda_\phi$, $\lambda_\alpha$, $\alpharb$, $\tau$, \textcolor{blue}{fixed temperature $\tilde{\alpha}$}}
$\bar{\phi}_1 \gets \phi_1,\ \bar{\phi_2} \gets \phi_2,\ \rbar_0 \gets 0,\ \mathcal{D} \gets \emptyset$ \;
\For{time step $t=0,1,2,\cdots$}{
$a_t \sim \pi_{\theta_t} (\cdot \mid s_t)$ \;
$s'_{t+1}, r_{t+1} \sim p(\cdot, \cdot \mid s_t, a_t)$ \;
$\rbar_{t+1} \doteq (1-\alpharb) \rbar_t + \alpharb\ r_{t+1}$ \;
\lIf{$s'_{t+1}$ is terminal}{$\tilde{s'}_{t+1} \sim d_0(\cdot)$} \lElse{$\tilde{s'}_{t+1} \doteq s'_{t+1}$}
$\mathcal{D} \gets \mathcal{D}\ \cup \{(s_t, a_t, r_{t+1}, \tilde{s'}_{t+1})\}$ \;
$\phi_{i,t+1} \doteq \phi_{i,t} - \lambda_\phi \nabla_{\phi_i} J_t (\phi_i)$, for $i \in \{1,2\}$ \;
\textcolor{red}{$\theta_{t+1} \doteq \theta_t + \lambda_\theta \nabla_\theta J_t (\theta)$} \;
\textcolor{red}{$\alpha_{t+1} \doteq \alpha_t - \lambda_\alpha \nabla_\alpha J_t (\alpha)$} \;
\textcolor{blue}{$\theta_{t+1} \doteq \theta_t + \lambda_\theta \nabla_\theta J_t^{\tilde{\alpha}} (\theta)$} \;
\textcolor{blue}{$\alpha_{t+1} \doteq \alpha_t - \hat{\lambda}_{\alpha,t} \nabla_\alpha J_t (\alpha)$} \;
$\bar{\phi}_{i,t+1} \doteq (1-\tau)\ \bar{\phi}_{i,t} + \tau \ \phi_{i,t}$, for $i \in \{1,2\}$
}
\caption{\textcolor{red}{Continuing SAC} or \textcolor{blue}{Continuing SAC with $\tilde{\alpha}$-toggle Intervention}}
\label{alg:CSAC}
\end{algorithm}
\vspace{-2ex}
\end{minipage}
\end{wrapfigure}

\begin{figure}[t]
    \centering
    \begin{minipage}{.48\linewidth}
        \centering
        \includegraphics[width=1\linewidth]{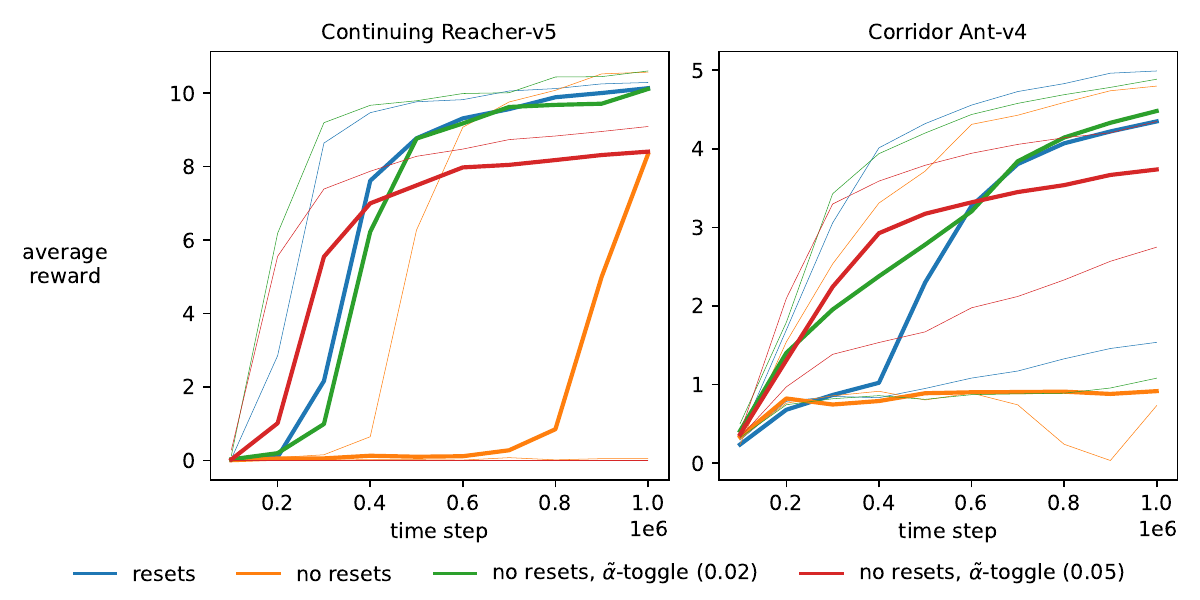}
        \caption{The $\tilde{\alpha}$-toggle intervention can recover the performance lost due to not using resets with the same value of $\tilde{\alpha}$ for both tasks. Dynamically adjusting the entropy of the policy via the entropy term in $J (\theta)$ is a viable alternative for improving exploration.}
        \label{fig:alpha_toggle}
    \end{minipage}
    \hfill
    \begin{minipage}{.48\linewidth}
        \centering
        \begin{minipage}{.49\linewidth}
            \includegraphics[width=1\linewidth]{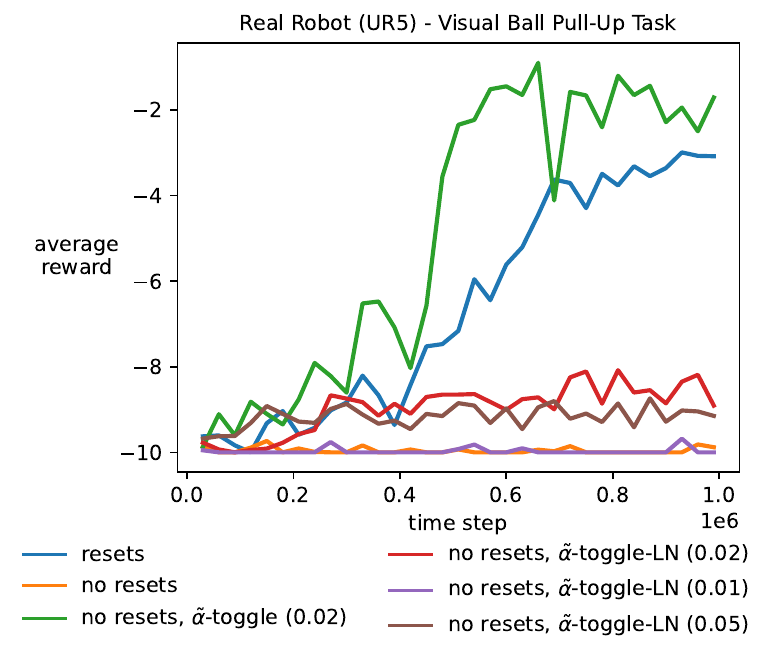}
        \end{minipage}
        \hfill
        \begin{minipage}{.49\linewidth}
            \includegraphics[width=1\linewidth]{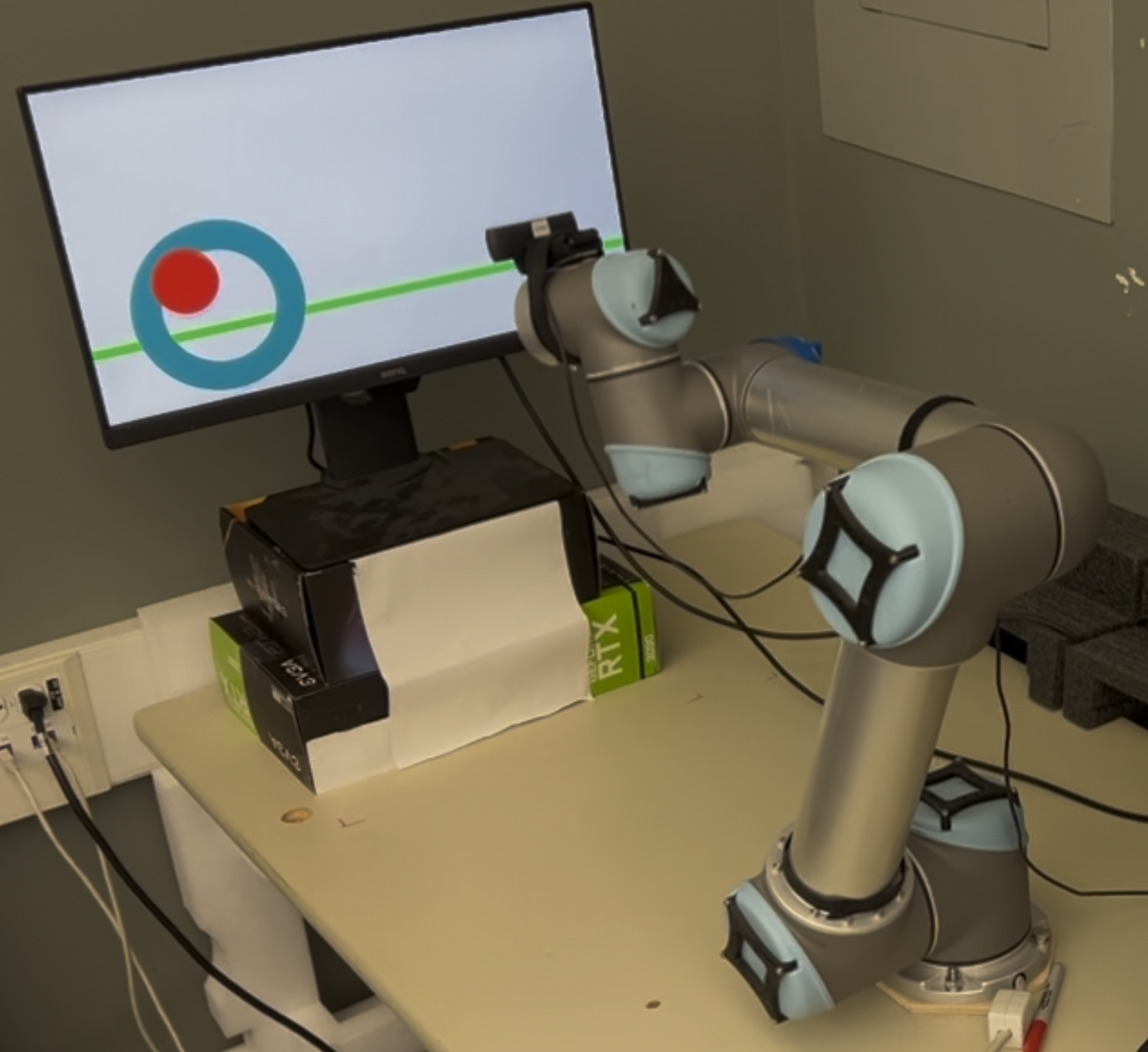}
        \end{minipage}
        \caption{Contrary to the results on simulated tasks, layer norm (omitted) and $\tilde{\alpha}$-toggle-LN fail to learn on the real-robot task. The $\tilde{\alpha}$-toggle intervention learns the task without resets with the same value of $\tilde{\alpha}=0.02$ that performed well in simulation tasks.}
        \label{fig:ur5_final}
    \end{minipage}
\end{figure}

\textbf{Can the entropy term of the policy objective counter the poor exploration caused by a high-variance Q function?}
The results from Figure \ref{fig:norms_critic} suggested that maximizing a Q function with high variance across states can result in poor exploration when learning without resets.
In addition to the $\min_{j \in \{1,2\}} Q_{\phi_{j,t}} (s_i, a_i)$ term, the objective $J (\theta)$ includes an entropy term $-\alpha_t \log \pi_{\theta_t} (a_i \mid s_i)$ with the auto-tuned temperature parameter $\alpha_t$.
Since $\alpha>0$, maximizing the entropy term increases the entropy of the policy, which increases the variance of the sampled actions, leading to better exploration.
Instead of modifying the action-value term for improving exploration, we now turn our attention to the entropy term.
We study whether dynamically adjusting $\alpha$ based on performance during learning can counter the poor exploration caused by maximizing a high-variance Q function.
We observed that when the simulated robot remains stationary without moving its joints for long periods, the performance measured by the average reward $\rbar$ in continuing SAC trends worse over time.
Based on this, we propose to increase the entropy of the policy when performance trends worse or remains at a fixed value over time.
Specifically, we estimate the performance in non-overlapping periods of 100 steps each.
The performance for period $i$ is $\rbar_i^{\tilde{\alpha}}$ and is calculated at the end of period $i$ by averaging the rewards received in that period.
If $\rbar_j^{\tilde{\alpha}} \leq \rbar_{j-1}^{\tilde{\alpha}}$ with $j$ denoting the most recent complete period, we use a constant temperature of $\tilde{\alpha}$ in the policy objective instead of the auto-tuned temperature parameter $\alpha$ and pause the learning updates of $\alpha$.
Otherwise, $\alpha$ is used in policy objective and learning updates of $\alpha$ are resumed.
We pause the learning updates of $\alpha$ by using the step size $\hat{\lambda}_{\alpha,t} \doteq \begin{cases} 0 \quad \text{if } \rbar_j^{\tilde{\alpha}} \leq \rbar_{j-1}^{\tilde{\alpha}},\ \lambda_\alpha \quad \text{otherwise}, \end{cases}\!\!\!\!\!\!\!$ with $\lambda_\alpha=0.001$.
In addition, to reduce sensitivity of performance to the value of $\tilde{\alpha}$, we scale each term in $J (\theta)$ (Hansen et al. 2024, Hafner et al. 2023).
Specifically, after sampling a mini-batch $\mathcal{M}_t \doteq \{ (s_i \sim \mathcal{D}, a_i \sim \pi_{\theta_t} (\cdot \mid s_i)) \}_{i=1}^{256}$, we update a moving average $\eta_{Q,t} \doteq 0.99\ \eta_{Q,t-1} + 0.01 \max(P_{95,t} - P_{5,t}, 1)$ of the scale of the Q values with $\eta_{Q,-1}=1$ and the percentile function $P_{k,t} \doteq \mathrm{Percentile} (k, \{ \min_{j \in \{1,2\}} Q_{\phi_{j,t}} (s_i,a_i) \}_{(s_i,a_i) \sim \mathcal{M}_t})$.
The policy objective can then be defined as
\begin{equation}
    J_t^{\tilde{\alpha}} (\theta) \doteq \mathbb{E}_{(s_i,a_i) \sim \mathcal{M}_t} \left[ \min_{j \in \{1,2\}} Q_{\phi_{j,t}} (s_i,a_i) \ /\ \eta_{Q,t}\ - \hat{\alpha}_t \log \pi_{\theta_t} (a_i \mid s_i) \ /\  |\mathcal{A}| \right], \nonumber
\end{equation}
where $\hat{\alpha}_t \doteq \begin{cases} \tilde{\alpha} \quad \text{if } \rbar_j^{\tilde{\alpha}} \leq \rbar_{j-1}^{\tilde{\alpha}},\ \alpha_t \quad \text{otherwise} \end{cases}\!\!\!\!\!\!\!$.
The parameter $\alpha$ is learned using the objective $J_t (\alpha) \doteq \mathbb{E}_{(s_i,a_i) \sim \mathcal{M}_t} \left[ -\alpha_t (\log \pi_{\theta_t} (a_i \mid s_i) + \targetent) \right]$.
We refer to this intervention as \emph{$\tilde{\alpha}$-toggle}.
Algorithm \ref{alg:CSAC} lists the pseudocode for continuing SAC with (blue) and without (red) $\tilde{\alpha}$-toggle.
Figure \ref{fig:alpha_toggle} depicts the performance of using $\tilde{\alpha}$-toggle when learning without resets for two fixed values of $\tilde{\alpha}$.
As shown, $\tilde{\alpha}$-toggle can recover the performance lost due to not using resets with the same value of $\tilde{\alpha}=0.02$ for both tasks.
The result shows that, aside from normalizing the Q function, dynamically adjusting the entropy of the policy via the entropy term in $J (\theta)$ is an alternative approach for countering the poor exploration caused by a high-variance Q function.

\begin{figure}[t]
  \centering
  \includegraphics[width=1.0\textwidth]{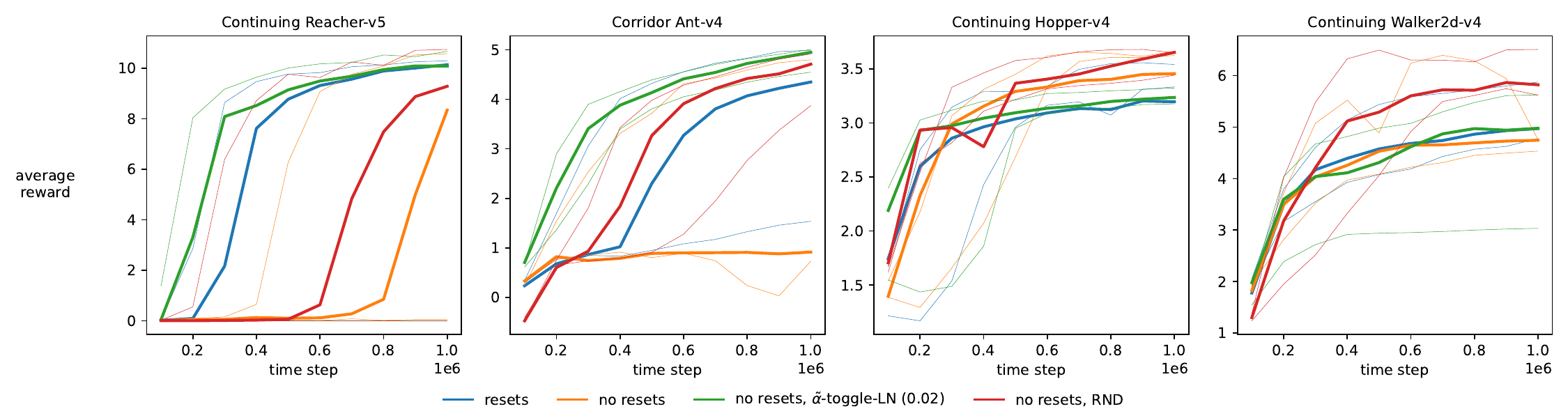}
  \caption{The $\tilde{\alpha}$-toggle-LN intervention with $\tilde{\alpha}=0.02$ performs well on all tasks. The $\tilde{\alpha}$-toggle-LN intervention learns faster than RND on continuing Reacher and Corridor Ant according to runs with median performance. None of the tested methods fail to learn on continuing Hopper and continuing Walker2d.}
  \label{fig:combined_final}
\end{figure}

Finally, we study the combination of $\tilde{\alpha}$-toggle and layer norm applied to actor and critic networks and refer to it as \emph{$\tilde{\alpha}$-toggle-LN}.
We ran $\tilde{\alpha}$-toggle-LN on four continuing Gym Mujoco environments and compared the results to the exploration method RND (Burda et al.\ 2019) in Figure \ref{fig:combined_final}.
We choose RND as a strong baseline method for exploration since it exceeded human performance in the hard-exploration environment of Montezuma's Revenge and can be implemented with minimal computational overhead.
We implemented the RND method in SAC following the official implementation in PPO and in the online manner as mentioned in Yang et al.\ (2024) to encourage exploration (Algorithm \ref{alg:RND}).
For each batch of samples used for updates, the intrinsic rewards were calculated, multiplied with the constant coefficient $c^{\text{int}}$, and added to the original rewards in the batch when calculating the Q function loss $J (\phi_i)$.
We tuned $c^{\text{int}}$ over values of 0.3, 1, and 3.
We kept track of the running mean and running variance of the observations in each batch and the calculated intrinsic rewards.
Using the running statistics, we normalized the observation inputs to the RND model and the calculated intrinsic rewards.
We used state-based resets in all experiments on continuing Hopper and continuing Walker2d since learning without state-based resets is out of our scope.
As seen in Figure \ref{fig:combined_final}, $\tilde{\alpha}$-toggle-LN performs well using $\tilde{\alpha}=0.02$ on all tasks.
Both $\tilde{\alpha}$-toggle-LN and RND are able to recover the performance lost due to not using resets.
However, $\tilde{\alpha}$-toggle-LN learns faster than RND on continuing Reacher and Corridor Ant according to runs with median performance.
None of the tested methods fail to learn on continuing Hopper and continuing Walker2d.

In addition, we tested the $\tilde{\alpha}$-toggle, layer norm, and $\tilde{\alpha}$-toggle-LN interventions on a vision task using a real UR5 robot arm with results shown in Figure \ref{fig:ur5_final} using one run for each experiment.
We use the SAC-Agent from the ReLoD learning system of Wang et al. (2023) in the Local-Only mode using a powerful workstation tethered to the robot,
 but replace the small convolutional network with a ResNet-18 model that is learned from scratch without pre-training and change SAC to the continuing SAC as described in section \ref{sec:continuing}.
We used the same value of $\tilde{\alpha}=0.02$ that performed well in simulation without tuning.
We refer to the task as \emph{ball pull-up}.
The goal of ball pull-up is to keep a 2D ball, which falls down with gravity, above the horizontal boundary line at the bottom of the monitor for as long as possible (see Appendix \ref{app:ball_pull_up}).
Despite multiple retries, using layer norm always resulted in emergency robot stops that halted the experiments early on.
Figure \ref{fig:ur5_final} shows that, contrary to the results on simulated tasks, layer norm (omitted) and $\tilde{\alpha}$-toggle-LN fail to learn on the real-robot task.
The $\tilde{\alpha}$-toggle intervention learns the task without resets.

\section{Related Work}

The continuing setup has been studied in the literature with solution methods that optimize either the average-reward or the discounted-reward criterion.
In solution methods that optimize the discounted-reward criterion, subtracting an estimate of the average reward from the TD error, also known as reward centering, has been studied by Naik et al.\ (2024) and the concurrent work of Wan et al.\ (2025).
Two common approaches for updating the average-reward estimate are as an exponential moving average of the observed rewards (simple), as used in our work, or based on the calculated TD errors (value based).
Naik et al.\ (2024) originally explored value-based centering of the rewards on Gym Mujoco tasks with PPO (Schulman et al.\ 2017) and on small-scale tasks with Q-learning (Watkins \& Dayan 1992) and DQN (Mnih et al.\ 2015).
Wan et al.\ (2025) found that despite the theoretical benefits of value-based centering in off-policy methods such as SAC, simple reward centering can perform just as well.
The continuing SAC algorithm presented in this paper is similar to the SAC algorithm with simple reward centering from Wan et al.\ (2025).
Our findings in Figure \ref{fig:sensit_alpha_gamma} that the performance of continuing SAC is less sensitive to the value of $\gamma$ than episodic SAC are in line with findings of Naik et al.\ (2024) and Wan et al.\ (2025) in that performance still declines in some tasks with large values of $\gamma$.

In contrast to their work, our paper studied possible mechanisms of failure and explored potential interventions for recovering the lost performance and improving exploration when learning without resets.
Our work further explicitly described how state-based reset transitions are treated in the continuing setup since the choice of the transition for adding a reset penalty to the reward can significantly affect performance.
Both Naik et al.\ (2024) and Wan et al.\ (2025) removed the time-based embodiment resets from the Gym Mujoco tasks in their study of the continuing setup with reward centering.
In our work, however, we explained in section \ref{sec:introduction} that episode terminations and embodiment resets are independent events and can occur in isolation.
Therefore, we studied the continuing SAC algorithm in section \ref{sec:continuing} while still using state-based and time-based embodiment resets.
Learning with the average-reward criterion has been applied to SAC by Hisaki and Ono (2024) by incorporating RVI Q-learning (Abounadi et al.\ 2001), and has in addition been applied to PPO by Ma et al.\ (2021) and to TRPO (Schulman et al.\ 2015) by Zhang and Ross (2021).

Works on reset-free RL aim to reduce the reliance on manual and hand-engineered resets when learning on real robots.
One approach for reset-free RL learns a separate reset policy (Eysenbach et al.\ 2017, Patil et al.\ 2024, Kim et al.\ 2022) to reverse the effects of the robot's actions on the environment since the previous reset.
Gupta et al.\ (2021) learn multiple related tasks in sequence such that the terminal state of one task can serve as the initial state of the next task.
Yang et al. (2025) encourage exploration by choosing goal states with high exploration value using a world model, which might not transfer to non-goal-reaching tasks, such as our real-robot vision task.
Other works that address exploration often encourage the agent to visit unseen states or unpredictable parts of the environment by adding a bonus to the received rewards.
For example, Tang et al.\ (2017) calculate the bonus based on state-visitation counts using a hash table.
The exploration bonus can also be calculated using the prediction error of a learned dynamics model (Pathak et al.\ 2017) or of a predictor network learning to match a random network as in RND (Burda et al.\ 2019).
In our work, we studied potential reasons for the poor performance of SAC when learning without resets and did not learn an additional reset policy.
Instead of adding an exploration bonus to the rewards, we studied simple computationally-cheap interventions targeted at the action-value and entropy terms of the policy objective of SAC.
Our results in Figure \ref{fig:r_bar_variance_ReacherNoReset-v5} are compatible with findings by Sharma et al.\ (2022) and Wan et al.\ (2025) that removing resets hinders the performance of SAC and reduces the diversity of visited states.

\section{Conclusion}

This paper explored the challenges of learning without episode terminations and time-based embodiment resets.
We addressed the first challenge of transforming from the episodic to the continuing task setup in section \ref{sec:continuing}.
We presented a continuing version of the SAC algorithm and showed that existing tasks can be transformed to the continuing setup via minimal modification of their reward function.
We studied the challenge of learning without time-based embodiment resets in section \ref{sec:continuing} and found that adequate exploration of the state space in SAC likely relies on a changing Q function during learning since a fixed Q function explored poorly.
In section \ref{sec:interventions}, we evaluated different approaches for improving exploration when learning without time-based resets.
Our results showed that, instead of large action magnitudes or saturated actions, high variance in the outputs of the critic network across states is the most likely cause of poor exploration.
We showed that exploration can be improved by using layer norm in the Q function or by increasing the entropy of the policy via dynamic adjustment of the temperature parameter.

\section*{Acknowledgments}
We are grateful to the Reinforcement Learning and Artificial Intelligence (RLAI) Laboratory, Alberta Machine Intelligence Institute (Amii), and the Canada CIFAR AI Chairs Program for their financial support, and to Kindred Inc. for their donation of the UR5 robot arm.
We also thank all of our colleagues in the Robot Lair group for the interesting and thoughtful discussions.

\newcommand{\hangin}{\goodbreak\hangindent=.35cm \noindent}
\section*{References}

\hangin
Abounadi, J., Bertsekas, D. P., Borkar, V. S. (2001). Learning algorithms for Markov decision processes with average cost. \textit{SIAM Journal on Control and Optimization, 40, 3}, 681–698.

\hangin
Ba, J. L., Kiros, J. R., Hinton, G. E. (2016). Layer Normalization. \textit{arXiv preprint arXiv:1607.06450}.

\hangin
Bjorck, J., Gomes, C. P., Weinberger, K. Q. (2022). Is high variance unavoidable in RL? A case study in continuous control. In \textit{International Conference on Learning Representations}.

\hangin
Brockman, G., Cheung, V., Pettersson, L., Schneider, J., Schulman, J., Tang, J., Zaremba, W. (2016). Openai gym. \textit{arXiv preprint arXiv:1606.01540}.

\hangin
Burda, Y., Edwards, H., Storkey, A., Klimov, O. (2019). Exploration by random network distillation. In \textit{7th International Conference on Learning Representations}.

\hangin
de Lazcano, R., Kallinteris, A., Tai, J. J., Lee, S. R., Terry, J. (2024). Gymnasium Robotics. URL \url{http://github.com/Farama-Foundation/Gymnasium-Robotics}

\hangin
Eysenbach, B., Gu, S., Ibarz, J., Levine, S. (2018). Leave no trace: Learning to reset for safe and autonomous reinforcement learning. In \textit{International Conference on Learning Representations}.

\hangin
Fujimoto, S., Hoof, H., Meger, D. (2018). Addressing function approximation error in actor-critic methods. In \textit{International Conference on Machine Learning}.

\hangin
Gupta, A., Yu, J., Zhao, T. Z., Kumar, V., Rovinsky, A., Xu, K., Devlin, T., Levine, S. (2021). Reset-free reinforcement learning via multi-task learning: Learning dexterous manipulation behaviors without human intervention. In \textit{2021 IEEE International Conference on Robotics and Automation}.

\hangin
Haarnoja, T., Zhou, A., Hartikainen, K., Tucker, G., Ha, S., Tan, J., Kumar, V., Zhu, H., Gupta, A., Abbeel, P., Levine, S. (2018). Soft actor-critic algorithms and applications. \textit{arXiv preprint arXiv:1812.05905}.

\hangin
Hafner, D., Pasukonis, J., Ba, J., Lillicrap, T. (2023). Mastering diverse domains through world models. \textit{arXiv preprint arXiv:2301.04104}.

\hangin
Hansen, N., Su, H., Wang, X. (2024). TD-MPC2: Scalable, robust world models for continuous control. In \textit{International Conference on Learning Representations}.

\hangin
Hisaki, Y., Ono, I. (2024). RVI-SAC: Average reward off-policy deep reinforcement learning. In \textit{Proceedings of the 41st International Conference on Machine Learning}.

\hangin
Kim, J., Park, J. H., Cho, D., Kim, H. J. (2022). Automating reinforcement learning with example-based resets. \textit{IEEE Robotics and Automation Letters, 7}, 3, 6606-6613.

\hangin
Kingma, D. P., Ba, J. (2014). Adam: A method for stochastic optimization. In \textit{3rd International Conference on Learning Representations}.

\hangin
Ma, X., Tang, X., Xia, L., Yang, J., Zhao, Q. (2021). Average-reward reinforcement learning with trust region methods. In \textit{Proceedings of the Thirtieth International Joint Conference on Artificial Intelligence}.

\hangin
Mnih, V., Kavukcuoglu, K., Silver, D., Rusu, A. A., Veness, J., Bellemare, M. G., Graves, A., Riedmiller, M., Fidjeland, A. K., Ostrovski, G., Petersen, S., Beattie, C., Sadik, A., Antonoglou, I., King, H., Kumaran, D., Wierstra, D., Legg, S., Hassabis, D. (2015). Human-level control through deep reinforcement learning. \textit{Nature, 518}, 7540, 529–533.

\hangin
Naik, A., Wan, Y., Tomar, M., Sutton, R. S. (2024). Reward centering. \textit{Reinforcement Learning Journal, 4}, 1995–2016.

\hangin
Pardo, F., Tavakoli, A., Levdik, V., Kormushev, P. (2018). Time limits in reinforcement learning. In \textit{Proceedings of the 35th International Conference on Machine Learning}.

\hangin
Pathak, D., Agrawal, P., Efros, A. A., Darrell, T. (2017). Curiosity-driven exploration by self-supervised prediction. In \textit{Proceedings of the 34th International Conference on Machine Learning}.

\hangin
Patil, D., Rajendran, J., Berseth, G., Chandar, S. (2024). Intelligent switching for reset-free RL. In \textit{International Conference on Learning Representations}.

\hangin
Schulman, J., Levine, S., Abbeel, P., Jordan, M., Moritz, P. (2015). Trust region policy optimization. In \textit{Proceedings of the 32nd International Conference on Machine Learning}.

\hangin
Schulman, J., Wolski, F., Dhariwal, P., Radford, A., Klimov, O. (2017). Proximal policy optimization algorithms. \textit{arXiv preprint arXiv:1707.06347}.

\hangin
Sharma, A., Xu, K., Sardana, N., Gupta, A., Hausman, K., Levine, S., and Finn, C. (2022). Autonomous reinforcement learning: Formalism and benchmarking. In \textit{International Conference on Learning Representations}.

\hangin
Sutton, R. S., Barto, A. G. (2018). \textit{Reinforcement learning: An introduction}. MIT Press.

\hangin
Tang, H., Houthooft, R., Foote, D., Stooke, A., Chen, X., Duan, Y., Schulman, J., De Turck, F., Abbeel, P. (2017). A study of count-based exploration for deep reinforcement learning. In \textit{Advances in Neural Information Processing Systems}.

\hangin
Tunyasuvunakool, S., Muldal, A., Doron, A., Liu, S., Bohez, S., Merel, J., Erez, T., Lillicrap, T., Heess, N., Tassa, Y. (2020). dm\_control: Software and tasks for continuous control. \textit{Software Impacts, 6}, 100022.

\hangin
van Hasselt, H. (2010). Double Q-learning. In \textit{Advances in Neural Information Processing Systems}.

\hangin
Vasan, G., Wang, Y., Shahriar, F., Bergstra, J., Jagersand, M., Mahmood, A. R. (2024). Revisiting sparse rewards for goal-reaching reinforcement learning. \textit{arXiv preprint arXiv:2407.00324}.

\hangin
Wan, Y., Korenkevych, D., Zhu, Z. (2025). An empirical study of deep reinforcement learning in continuing tasks. \textit{arXiv preprint arXiv:2501.06937}.

\hangin
Wang, Y., Vasan, G., Mahmood, A. R. (2023). Real-time reinforcement learning for vision-based robotics utilizing local and remote computers. In \textit{2023 IEEE International Conference on Robotics and Automation}.

\hangin
Watkins, C. J. C. H., Dayan, P. (1992). Q-learning. \textit{Machine Learning, 8}, 279-292.

\hangin
Yang, K., Tao, J., Lyu, J., Li, X. (2024). Exploration and anti-exploration with distributional random network distillation. In \textit{Proceedings of the 41st International Conference on Machine Learning}.

\hangin
Zhao, Y., Ding, Y., Wen, Y., Gu, S., Levine, S. (2025). Reset-free reinforcement learning with world models. \textit{Transactions on Machine Learning Research}.

\hangin
Zhang, Y., Ross, K. (2021). On-policy deep reinforcement learning for the average-reward criterion. In \textit{Proceedings of the 38th International Conference on Machine Learning}.

\clearpage

\appendix
\section{Appendix}

\subsection{Random Network Distillation (RND) Algorithm}

The RND method (Burda et al. 2019) uses a target network $f_{\bar{\psi}}: \mathbb{R}^{|\mathcal{S}|} \to \mathbb{R}^{256}$ and a predictor network $f_\psi: \mathbb{R}^{|\mathcal{S}|} \to \mathbb{R}^{256}$.
The parameters $\bar{\psi}$ are randomly initialized and fixed for the experiment, and parameters $\psi$ are learned.
For each sample in the mini-batch $\mathcal{M}_t \doteq \{ (s_i \sim \mathcal{D},\ a_i \sim \pi_{\theta_t} (\cdot \mid s_i),\ s'_i) \}_{i=1}^{256}$, RND calculates the intrinsic reward $r_i^\text{int} \doteq \frac{1}{2} \| f_{\psi_t} (s_i) - f_{\bar{\psi}} (s_i) \|_2^2$, which is then used to calculate the action-value objective
\begin{equation}
    J_t^\text{rnd} (\phi_i) \doteq \frac{1}{2} \mathbb{E}_{(s_i,a_i,r_i,s'_i) \sim \mathcal{D}} \left[ \left[ r_i + c^{\text{int}} r_i^{\text{int}} - \rbar_{t+1} + \gamma \left( \min_{j \in \{1, 2\}} Q_{\bar{\phi}_{j,t}} (s'_i,a'_i) - \alpha_t \log \pi_{\theta_t} (a'_i \mid s'_i) \right) - Q_{\phi_{i,t}} (s_i,a_i) \right]^2 \right], \nonumber
\end{equation}
with intrinsic reward coefficient $c^{\text{int}}$ and $a'_i \sim \pi_{\theta_t} (\cdot \mid s'_i)$.
To learn the parameters $\psi$, the loss is calculated for each sample as $l (\psi_t, s'_i) \doteq \frac{1}{|\mathcal{S}|} \|f_{\psi_t} (s'_i) - f_{\bar{\psi}} (s'_i) \|_2^2$, leading to the masked and scaled RND objective
\begin{equation}
    J_t (\psi) \doteq \frac{\sum_{i=1}^{256} m_i\ l (\psi_t, s'_i)}{\max (\sum_{i=1}^{256} m_i, 1)},
\end{equation}
with masks $m_i \sim \text{Bernoulli} (p^{\text{upd}})$ and update proportion hyper-parameter $p^{\text{upd}}$.
In addition, the inputs to the functions $f_\psi$ and $f_{\bar{\psi}}$ are normalized by subtracting the running mean and dividing by the running standard deviation for each state dimension separately.
The mean and standard deviation are calculated over samples of all mini-batches.
The intrinsic rewards are normalized similarly using their running mean and standard deviation.
The algorithm below lists the pseudocode for continuing SAC with RND.

\begin{algorithm}[H]
\DontPrintSemicolon
\KwIn{dynamics $p$, initial-state distribution $d_0$, weights $\theta$, $\phi_1$, $\phi_2$, $\psi$, $\bar{\psi}$, step sizes $\lambda_\theta$, $\lambda_\phi$, $\lambda_\alpha$, $\alpharb$, $\tau$, $\lambda_\psi$}
$\bar{\phi}_1 \gets \phi_1,\ \bar{\phi_2} \gets \phi_2,\ \rbar_0 \gets 0,\ \mathcal{D} \gets \emptyset$ \;
\For{time step $t=0,1,2,\cdots$}{
$a_t \sim \pi_{\theta_t} (\cdot \mid s_t)$ \;
$s'_{t+1}, r_{t+1} \sim p(\cdot, \cdot \mid s_t, a_t)$ \;
$\rbar_{t+1} \doteq (1-\alpharb) \rbar_t + \alpharb\ r_{t+1}$ \;
\lIf{$s'_{t+1}$ is terminal}{$\tilde{s'}_{t+1} \sim d_0(\cdot)$} \lElse{$\tilde{s'}_{t+1} \doteq s'_{t+1}$}
$\mathcal{D} \gets \mathcal{D}\ \cup \{(s_t, a_t, r_{t+1}, \tilde{s'}_{t+1})\}$ \;
$\phi_{i,t+1} \doteq \phi_{i,t} - \lambda_\phi \nabla_{\phi_i} J_t^\text{rnd} (\phi_i)$, for $i \in \{1,2\}$ \;
$\psi_{t+1} \doteq \psi_t - \lambda_\psi \nabla_\psi J_t (\psi)$ \;
$\theta_{t+1} \doteq \theta_t + \lambda_\theta \nabla_\theta J_t (\theta)$ \;
$\alpha_{t+1} \doteq \alpha_t - \lambda_\alpha \nabla_\alpha J_t (\alpha)$ \;
$\bar{\phi}_{i,t+1} \doteq (1-\tau)\ \bar{\phi}_{i,t} + \tau \ \phi_{i,t}$, for $i \in \{1,2\}$
}
\caption{Continuing SAC with Random Network Distillation (RND)}
\label{alg:RND}
\end{algorithm}

\clearpage
\subsection{Hyper-parameters}

\begin{table}[h]
\centering
\textbf{Continuing SAC Hyper-Parameters}\\
\begin{tabular}{ll}
\toprule
\textbf{Hyper-parameter} & \textbf{Value} \\
\midrule
\# of hidden layers & 2 \\
\# of hidden units & 256 \\
activation & ReLU \\
optimizer & Adam (Kingma \& Ba 2014) \\
replay buffer size & $10^6$ \\
mini-batch size & 256 \\
discount rate $\gamma$ & 0.99 \\
target entropy $\bar{\mathcal{H}}$ & $-|\mathcal{A}|$ \\
actor step size $\lambda_\theta$ & $3 \times 10^{-4}$ \\
critic step size $\lambda_\phi$ & $1 \times 10^{-4}$ \\
temperature step size $\lambda_\alpha$ & $1 \times 10^{-4}$ \\
target Q step size $\tau$ & 0.005 \\
average reward step size $\alpharb$ & $3 \times 10^{-4}$ \\
\bottomrule
\end{tabular}
\end{table}

\begin{table}[h]
\centering
\textbf{RND Hyper-Parameters}\\
\begin{tabular}{ll}
\toprule
\textbf{Hyper-parameter} & \textbf{Value} \\
\midrule
\# of hidden layers (predictor) & 4 \\
\# of hidden layers (target) & 2 \\
\# of hidden units & 256 \\
activation & ReLU \\
\# of outputs & 256 \\
RND step size $\lambda_\psi$ & $3 \times 10^{-4}$ \\
update proportion $p^{\text{upd}}$ & $0.25$ \\
intrinsic reward coefficient $c^{\text{int}}$ & $\{0.3, 1, 3\}$ \\
\bottomrule
\end{tabular}
\end{table}

\clearpage
\subsection{Additional Experiments}

\subsubsection{Continuing SAC on DMControl Tasks}

\begin{figure}[ht]
  \centering
  \begin{minipage}{\linewidth}
      \centering
      \includegraphics[width=1.0\textwidth]{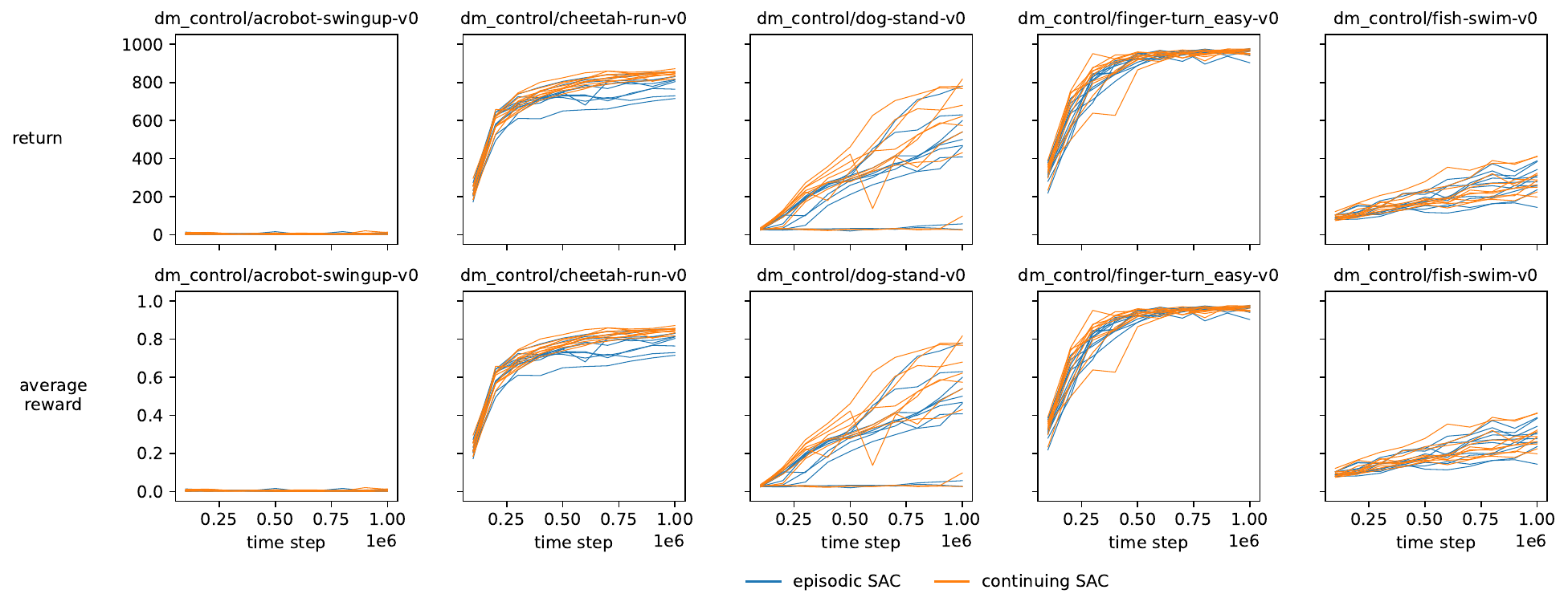}
  \end{minipage}
  \begin{minipage}{\linewidth}
      \centering
      \includegraphics[width=1.0\textwidth]{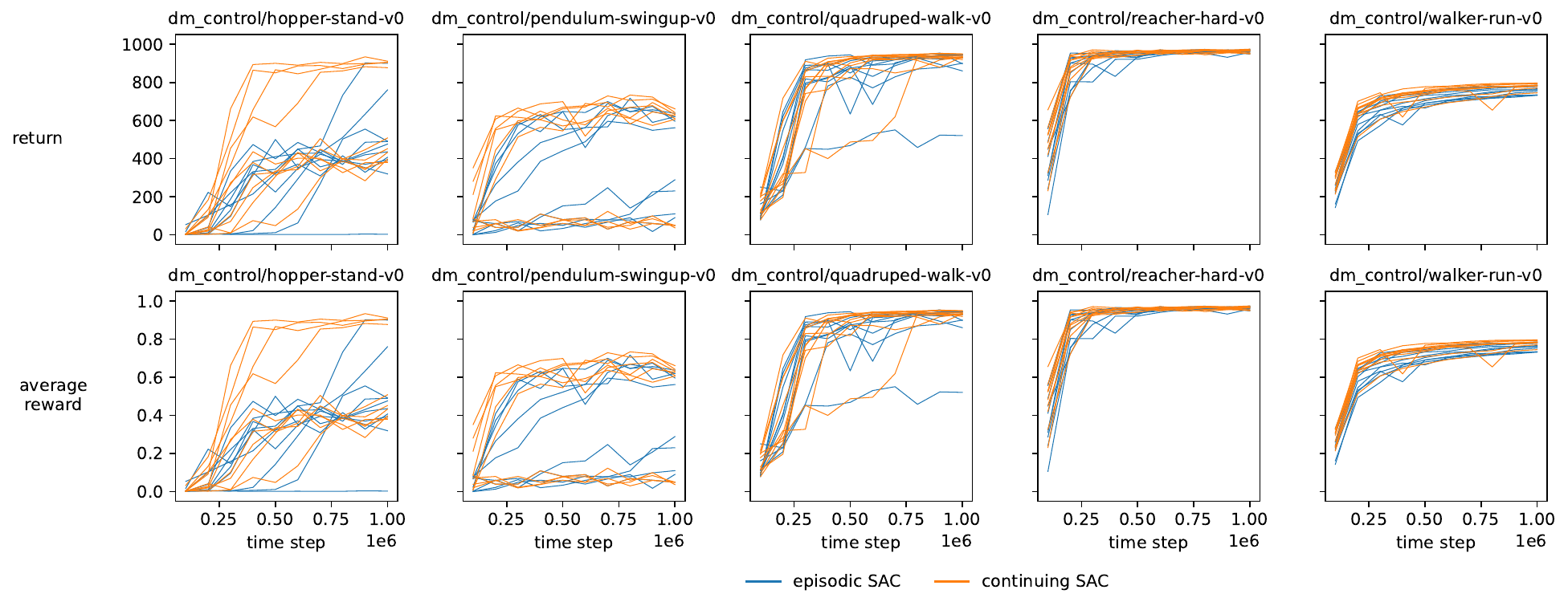}
      \caption{Continuing SAC (orange) performs similarly to episodic SAC (blue) in the tested DMControl tasks without any modifications to the reward functions of tasks.}
      \label{fig:return_vs_rbar_term_reset_bootstrap_dm_control}
  \end{minipage}
\end{figure}

\begin{figure}[ht]
  \centering
  \begin{minipage}{\linewidth}
      \centering
      \includegraphics[width=1.0\textwidth]{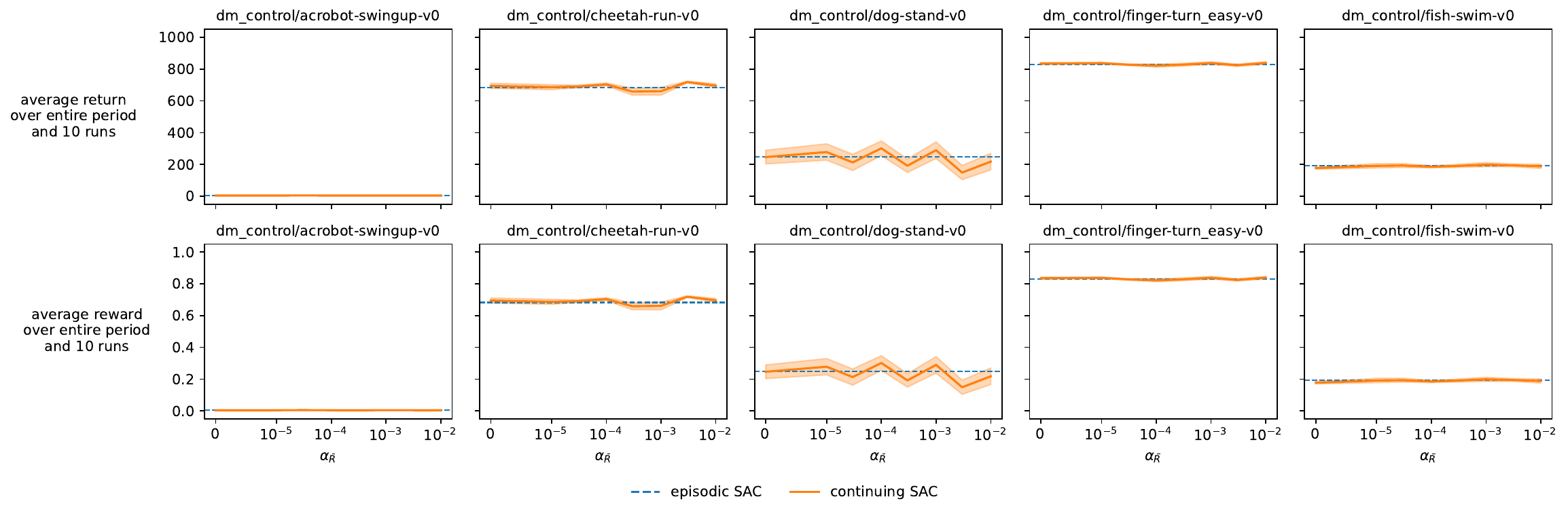}
  \end{minipage}
  \begin{minipage}{\linewidth}
      \centering
      \includegraphics[width=1.0\textwidth]{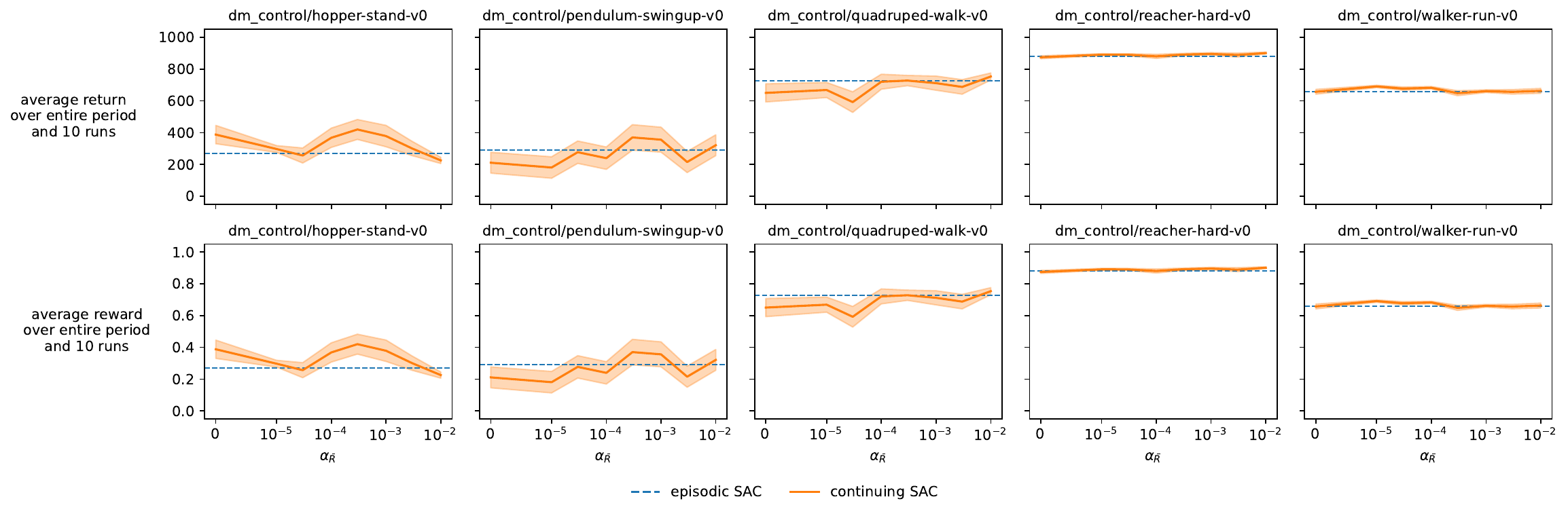}
      \caption{Continuing SAC (orange) is not sensitive to the value of the step size of average reward $\alpha_{\rbar}$.}
      \label{fig:sensit_term_reset_bootstrap_dm_control}
  \end{minipage}
\end{figure}

\clearpage
\subsubsection{Continuing SAC with Different Penalty Values for State-Based Resets}

\begin{figure}[ht]
  \centering
  \begin{minipage}{\linewidth}
      \centering
      \includegraphics[width=0.88\textwidth]{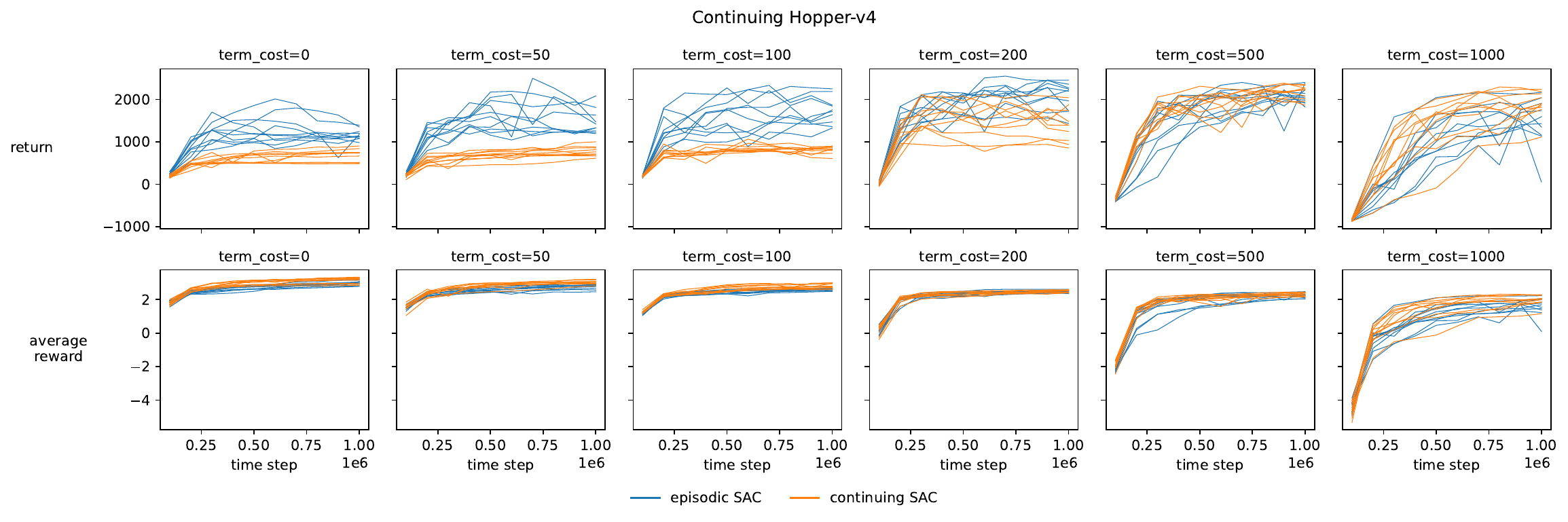}
  \end{minipage}
  \begin{minipage}{\linewidth}
      \centering
      \includegraphics[width=0.88\textwidth]{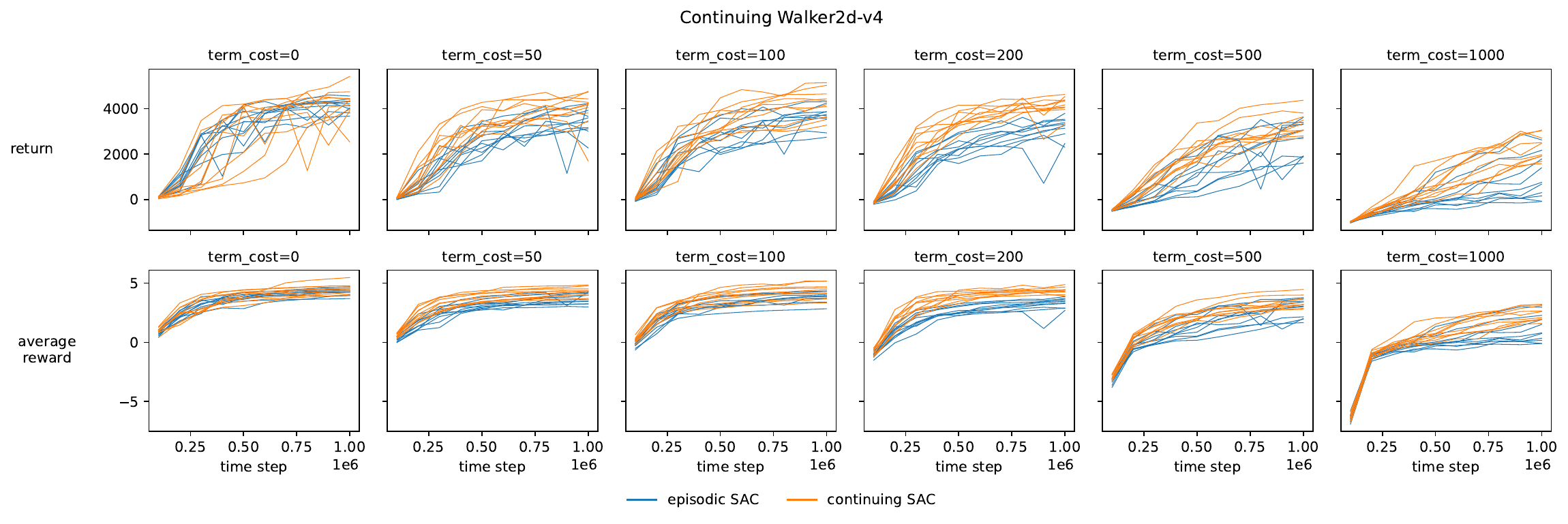}
  \end{minipage}
  \begin{minipage}{\linewidth}
      \centering
      \includegraphics[width=0.88\textwidth]{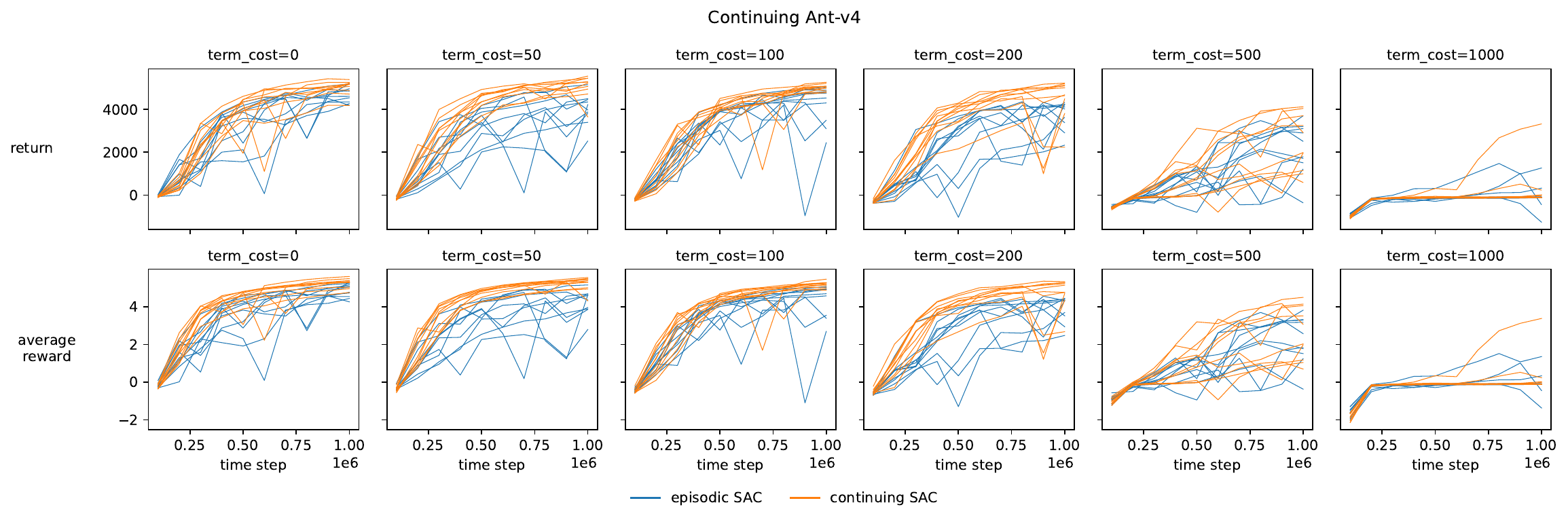}
  \end{minipage}
  \begin{minipage}{\linewidth}
      \centering
      \includegraphics[width=0.88\textwidth]{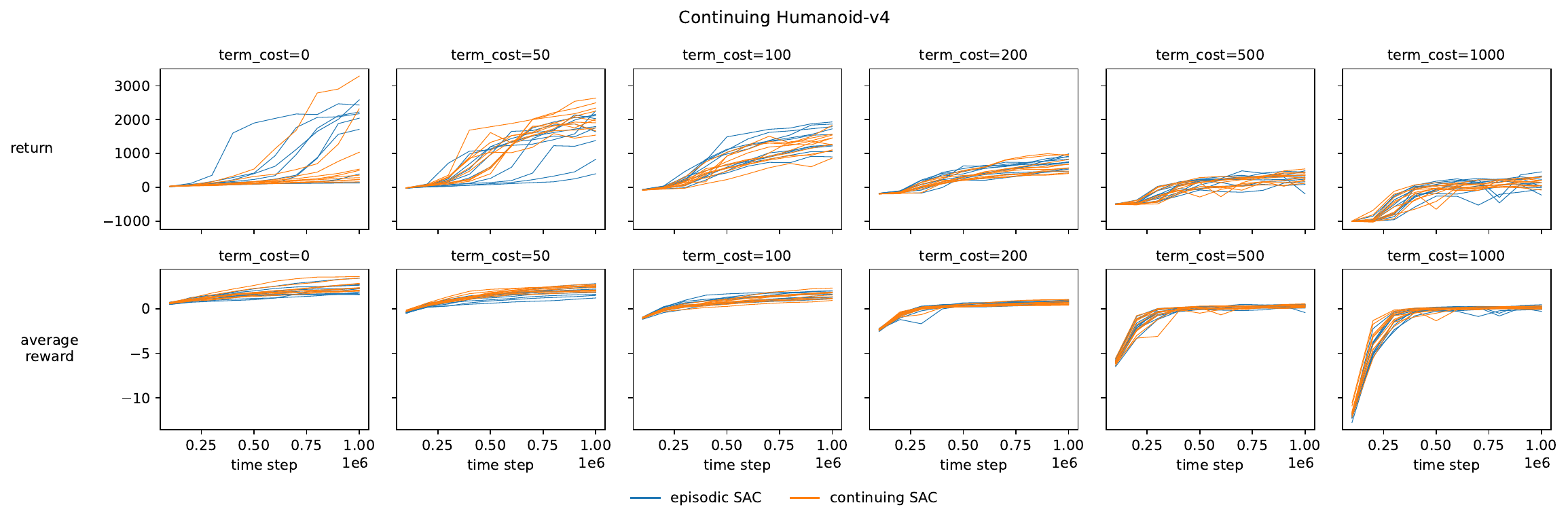}
      \caption{Performance of continuing SAC (orange) compared with episodic SAC (blue) using different penalty values (term\_cost) for state-based resets. Plotted are the results measured using the modified reward function including the penalty, not the original reward function of the task.}
  \end{minipage}
  \label{fig:penalty_values}
\end{figure}

\clearpage
\subsubsection{Periodic Resetting of Actor and Critic Networks}

We tested resetting the actor and critic network parameters excluding the temperature parameter $\alpha$, referred to as \emph{network reset}, every 5000, 10,000, and 20,000 time steps when learning without resets.
We also tested two additional variations of network reset.
With every network reset, \emph{Network reset-$\alpha$} also resets the value of the temperature parameter to a random value from $\mathcal{U}[0.01, 1]$ and does not auto-tune $\alpha$.
With every network reset, \emph{Network reset-$\mathcal{H}$} also resets the value of target entropy $\bar{\mathcal{H}}$ to a random value from $\mathcal{U}[-|\mathcal{A}|, 0.675\ |\mathcal{A}|]$ and auto-tunes $\alpha$.
The $-|\mathcal{A}|$ value is SAC's default.
The $0.675\ |\mathcal{A}|$ value is close to the empirically-observed maximum value for the entropy of the policy.

\begin{figure}[ht]
  \centering
  \begin{minipage}{\linewidth}
      \centering
      \includegraphics[width=1.0\textwidth]{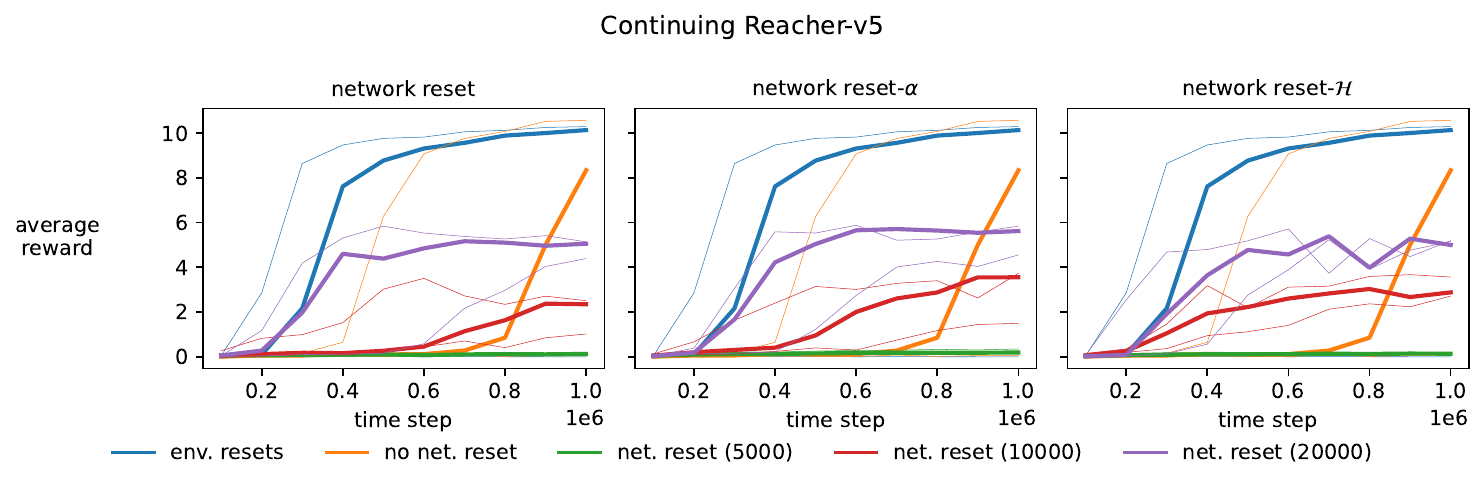}
  \end{minipage}
  \begin{minipage}{\linewidth}
      \centering
      \includegraphics[width=1.0\textwidth]{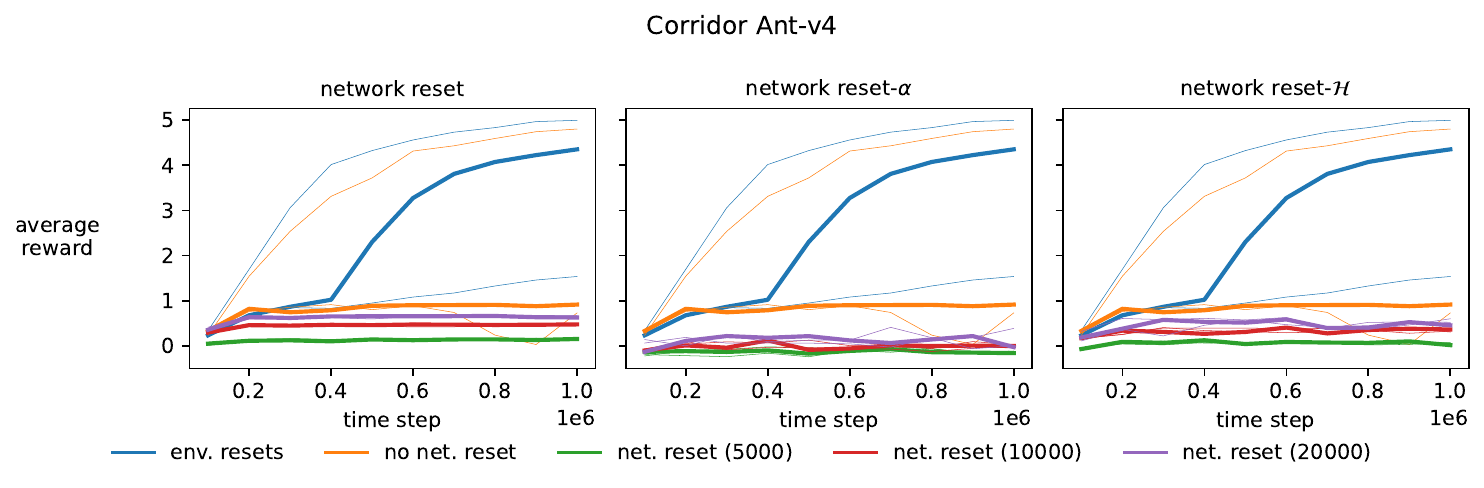}
      \caption{Resetting the actor and critic networks periodically does not fully recover the performance lost due to not using resets. Resetting with periods of every 10,000 and 20,000 steps help initially in continuing Reacher but plateau and prevent higher performance.}
  \end{minipage}
  \label{fig:no_reset_agent_bias_resets_agent}
\end{figure}

\clearpage
\subsubsection{The $\tilde{\alpha}$-toggle Intervention With Different Periods}

\begin{figure}[ht]
  \centering
  \begin{minipage}{\linewidth}
      \centering
      \includegraphics[width=1.0\textwidth]{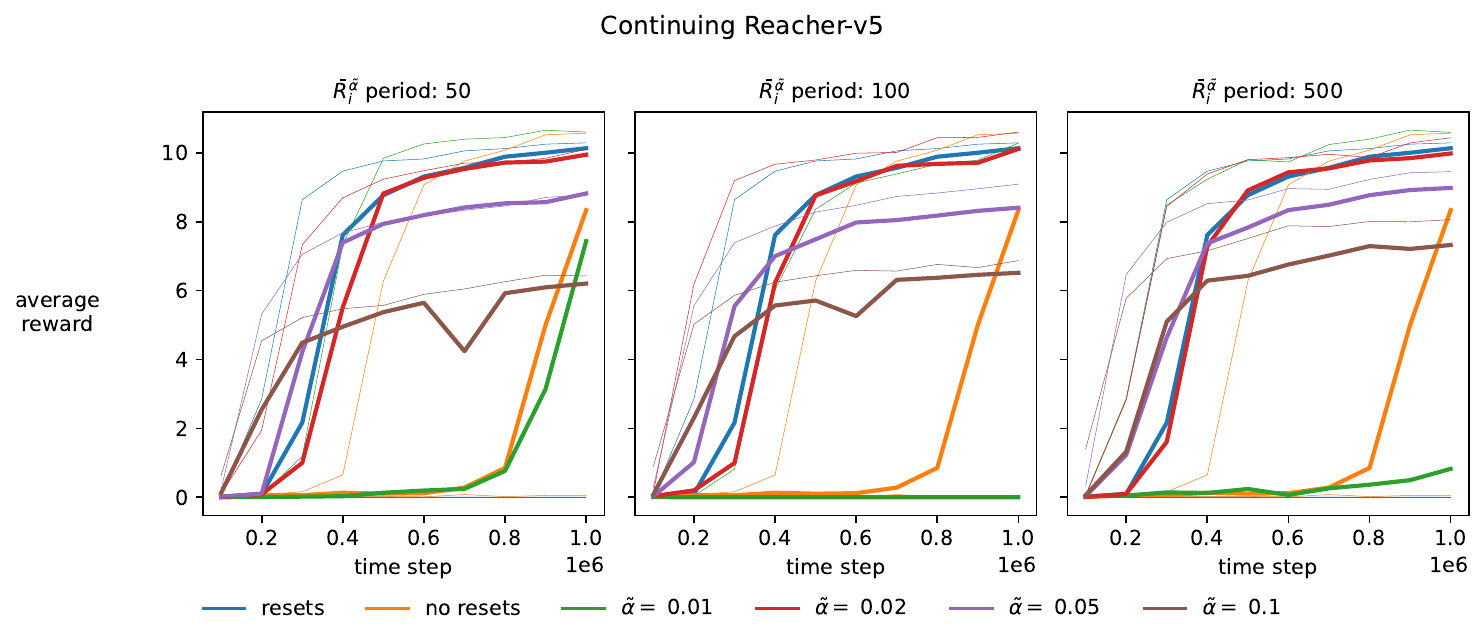}
  \end{minipage}
  \begin{minipage}{\linewidth}
      \centering
      \includegraphics[width=1.0\textwidth]{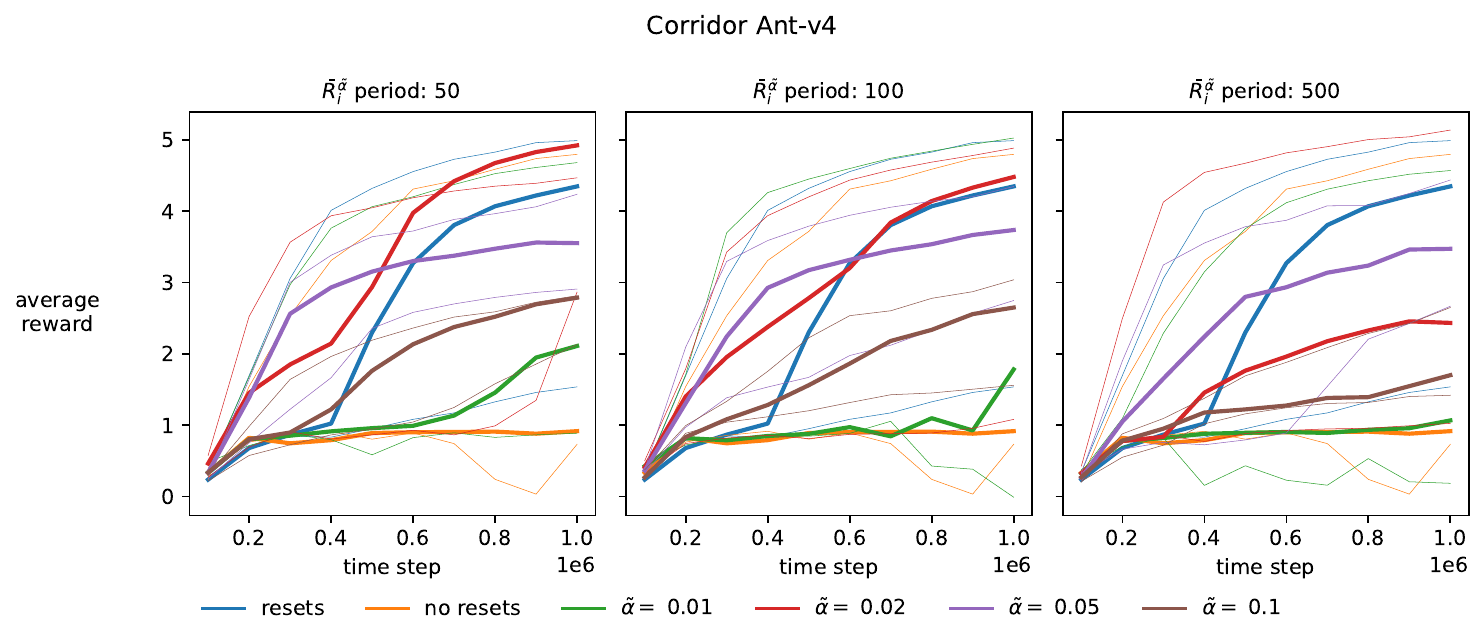}
      \caption{Performance is comparable between different periods. Curves with $\tilde{\alpha}$ values use the $\tilde{\alpha}$-toggle intervention.}
  \end{minipage}
  \label{fig:no_reset_adapt_te_high_alpha}
\end{figure}

\clearpage
\subsubsection{Applying Either Layer Normalization or PNorm to Actor and Critic Networks Simultaneously}

\begin{figure}[ht]
    \centering
    \begin{minipage}{.48\linewidth}
        \centering
        \includegraphics[width=1\linewidth]{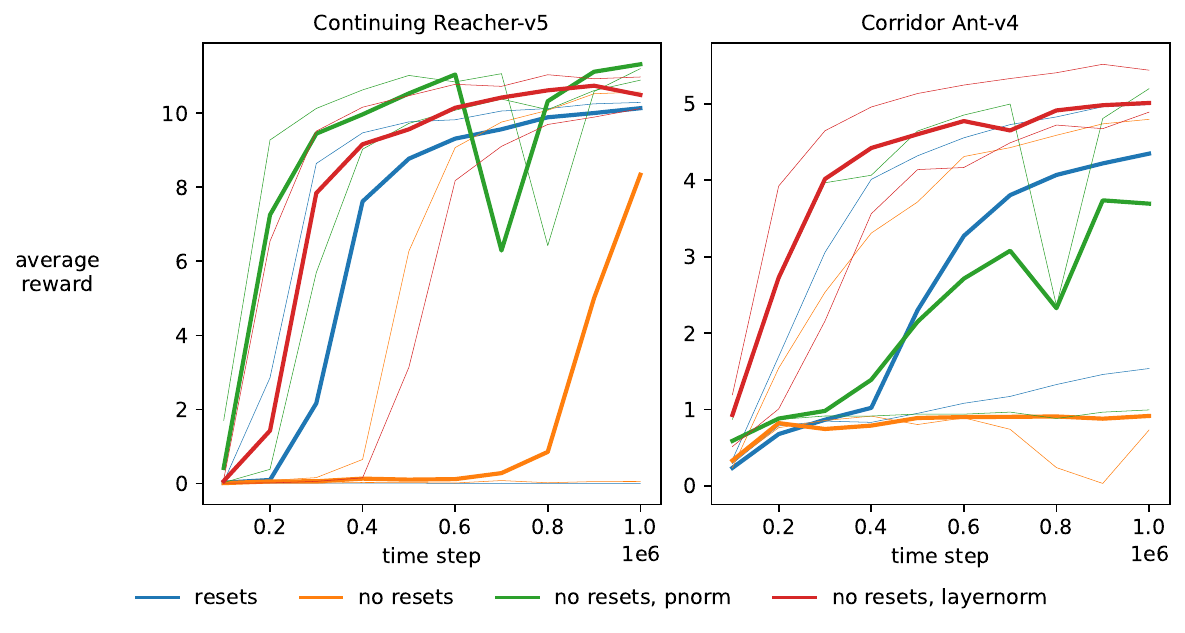}
        \caption{Applying either layer norm or pnorm to both networks is comparable to applying to the critic network only seen in Figure \ref{fig:norms_critic}. Performance dips momentarily toward the end when pnorm is applied to both networks.}
        \label{fig:norms}
    \end{minipage}
    \hfill
    \begin{minipage}{.48\linewidth}
        \centering
    \end{minipage}
\end{figure}

\clearpage
\subsubsection{Using Normalization or $\tilde{\alpha}$-toggle When Learning With Resets}
\begin{figure}[ht]
    \centering
    \begin{minipage}{.48\linewidth}
        \centering
        \includegraphics[width=1\linewidth]{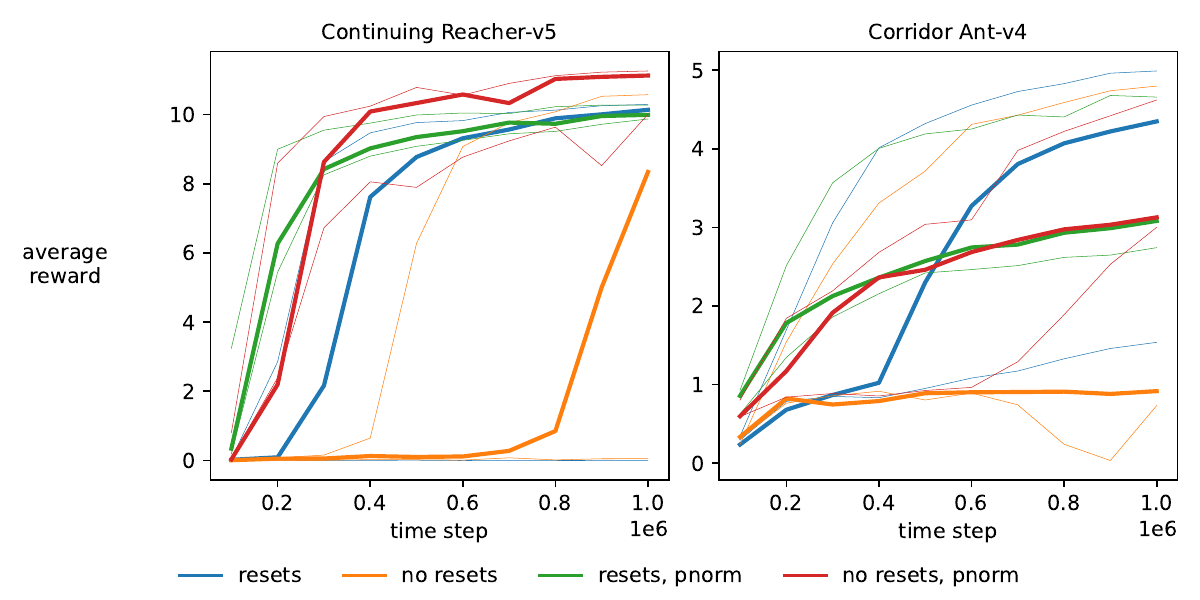}
        \caption{With resets, using pnorm may result in slightly faster learning in the beginning but similar or worse asymptotic performance according to the median runs.}
        \label{fig:norms_critic_pnorm}
    \end{minipage}
    \hfill
    \begin{minipage}{.48\linewidth}
        \centering
        \includegraphics[width=1\linewidth]{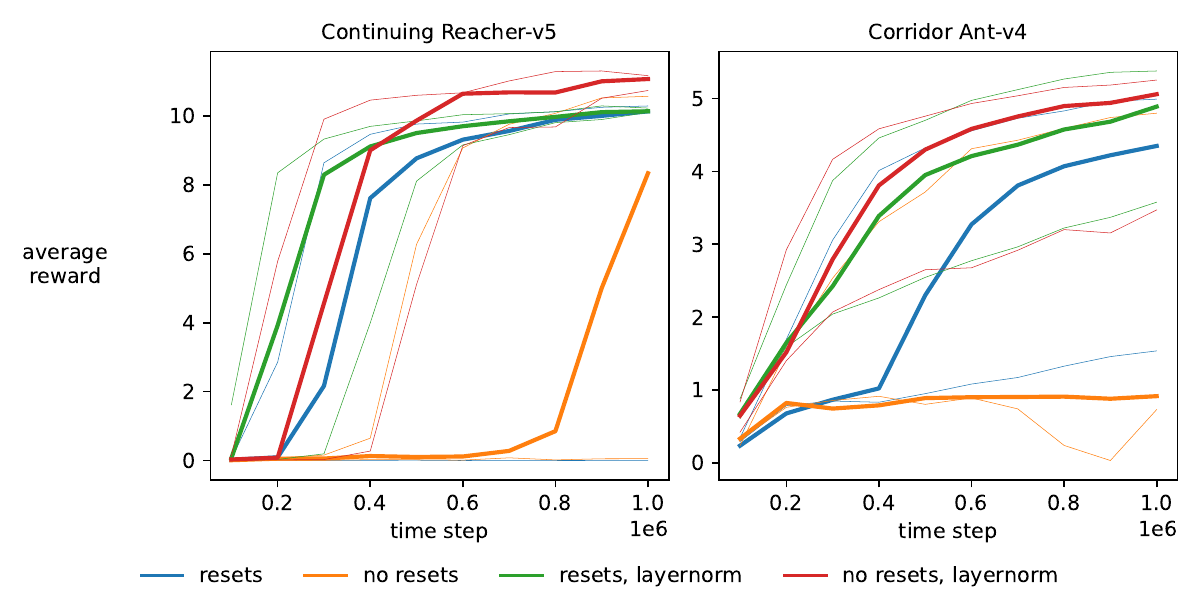}
        \caption{With resets, using layer norm may result in slightly faster learning in the beginning according to the median runs.}
        \label{fig:norms_critic_layernorm}
    \end{minipage}
\end{figure}

\begin{figure}[ht]
    \centering
    \begin{minipage}{.48\linewidth}
        \centering
        \includegraphics[width=1\linewidth]{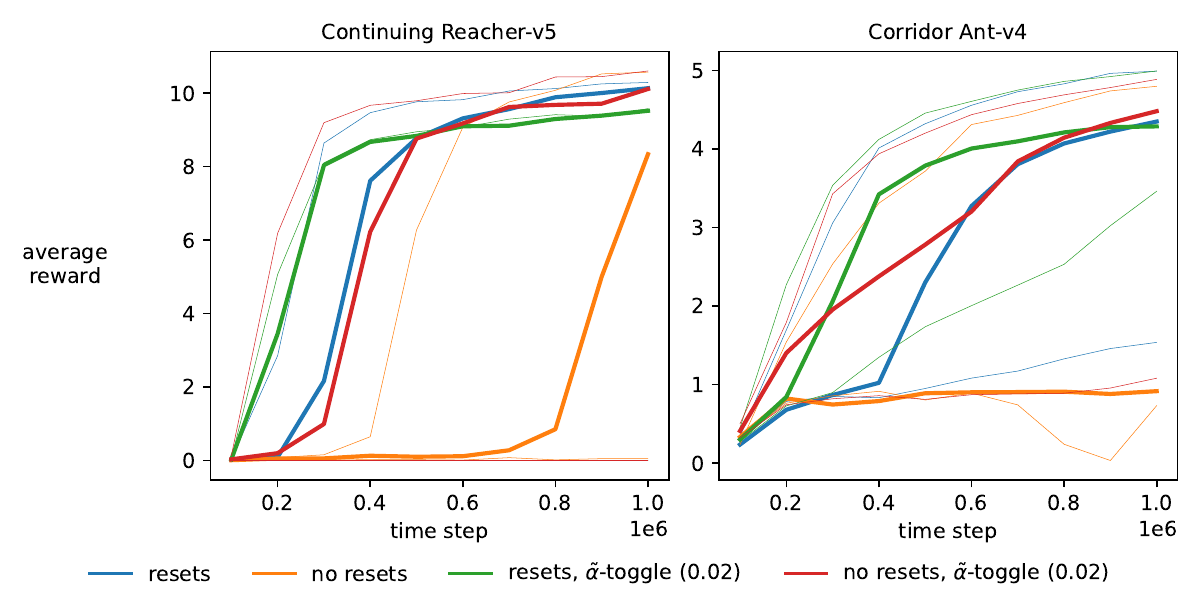}
        \caption{The $\tilde{\alpha}$-toggle intervention ($\tilde{\alpha}=0.02$) may slightly increase learning speed when learning with resets according to the median runs.}
        \label{fig:alpha_toggle_scale_q_ha.02}
    \end{minipage}
    \hfill
    \begin{minipage}{.48\linewidth}
        \centering
        \includegraphics[width=1\linewidth]{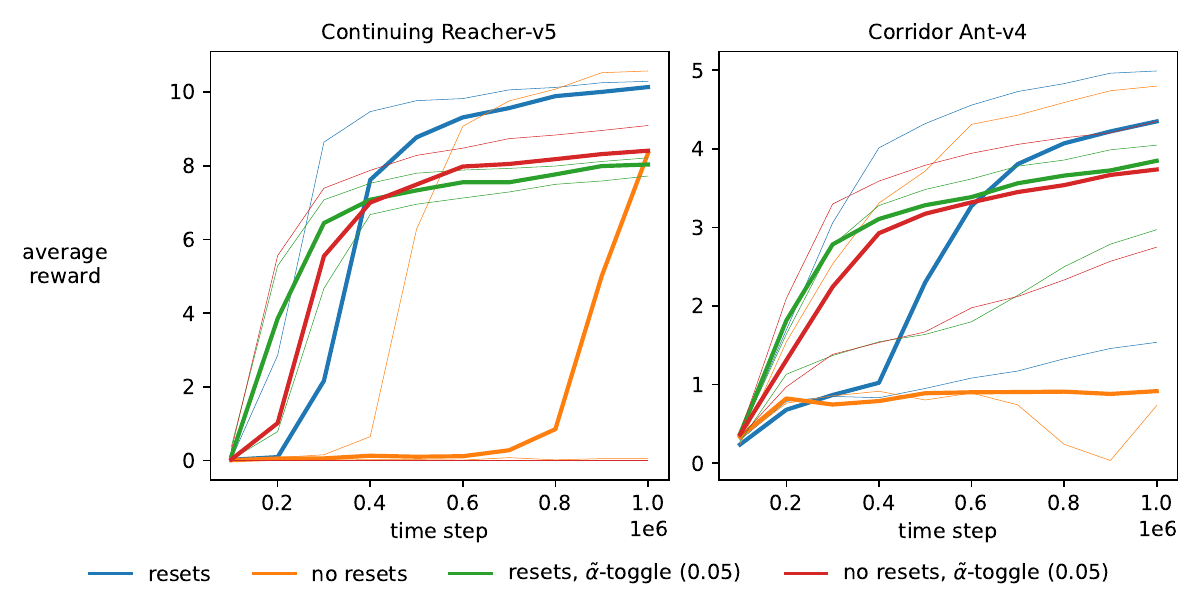}
        \caption{The $\tilde{\alpha}$-toggle intervention ($\tilde{\alpha}=0.05$) may slightly increase learning speed when learning with resets according to the median runs.}
        \label{fig:alpha_toggle_scale_q_ha.05}
    \end{minipage}
\end{figure}

\clearpage
\subsubsection{Different Values of $\tilde{\alpha}$ on the Visual Ball Pull-Up Task}
\begin{figure}[ht]
    \centering
    \includegraphics[width=1\linewidth]{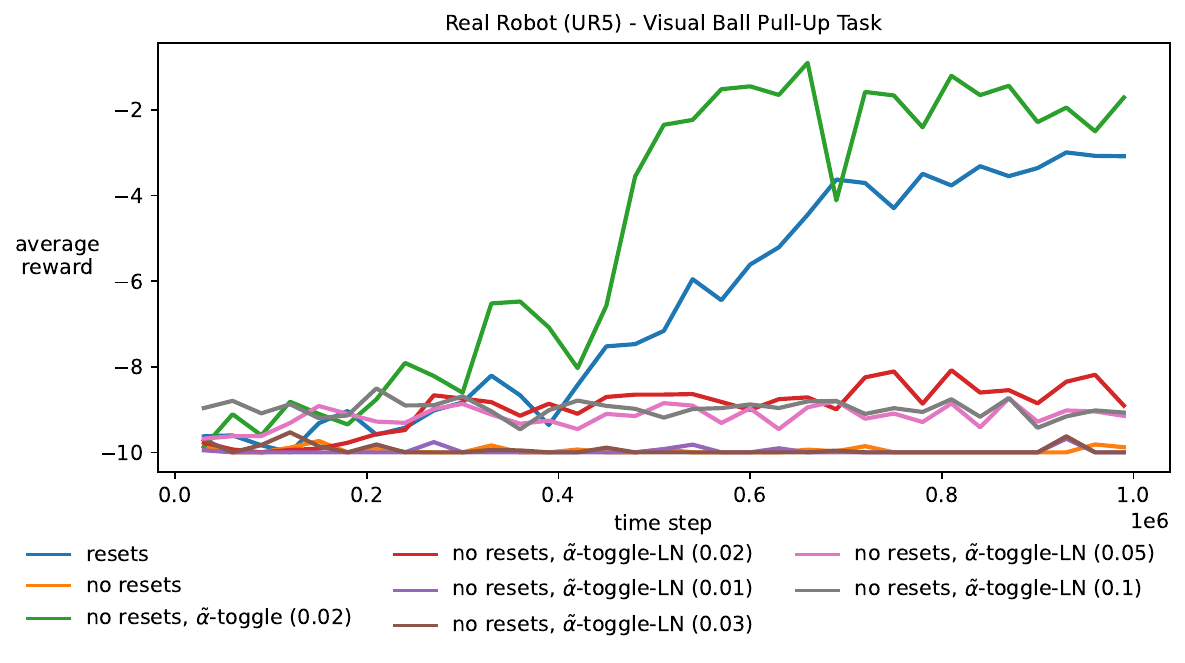}
    \caption{Only the $\tilde{\alpha}$-toggle intervention is able to learn the real-robot task without resets. The $\tilde{\alpha}$-toggle-LN intervention fails to learn with all tested values of $\tilde{\alpha}$.}
    \label{fig:ur5_final_scale_q_vary_alpha_all}
\end{figure}

\clearpage
\subsubsection{Comparison of Different Interventions on Sparse-Reward Tasks}
\begin{figure}[ht]
  \centering
  \begin{minipage}{\linewidth}
      \centering
      \includegraphics[width=1.0\textwidth]{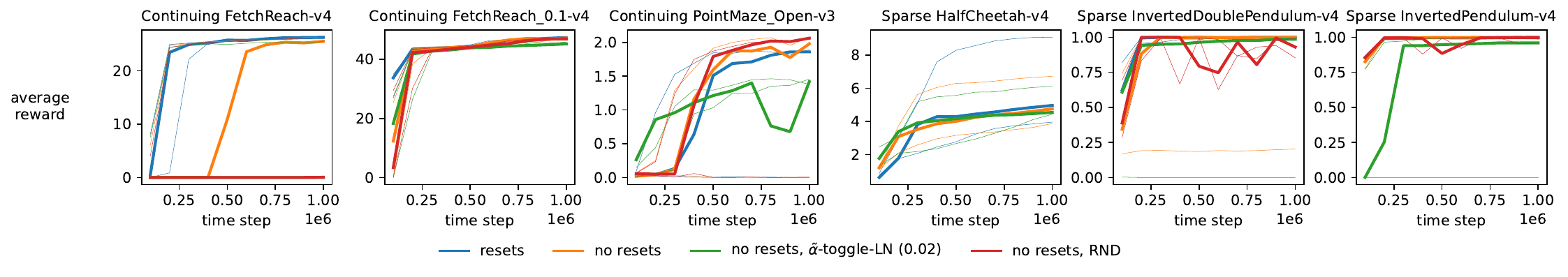}
  \end{minipage}
  \begin{minipage}{\linewidth}
      \centering
      \includegraphics[width=1.0\textwidth]{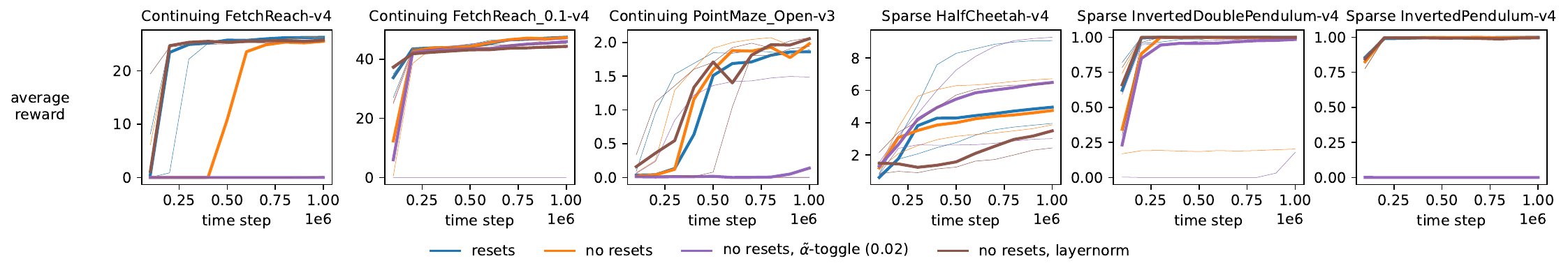}
      \caption{None of the tested sparse-reward tasks fail to learn without resets. The performance of the $\tilde{\alpha}$-toggle-LN intervention with $\tilde{\alpha}=0.02$ is comparable to the performance of learning with resets on all tasks except continuing FetchReach and continuing PointMaze. Only the layer norm intervention can recover the lost performance when learning without resets in continuing FetchReach.}
      \label{fig:sparse_tasks}
  \end{minipage}
\end{figure}

We compared the performance of different interventions for learning without resets on five additional sparse-reward tasks.
Figure \ref{fig:sparse_tasks} shows the learning curves.
After sorting the 10 independent runs of each experiment in ascending order of average performance over the entire period, we show the median (5th) run with a thick line and the first and last runs with thin lines.
The \emph{continuing FetchReach} and \emph{continuing PointMaze} tasks were adapted from the FetchReach-v4 and PointMaze\_Open\_Diverse\_G-v3 tasks in Gymnasium Robotics (de Lazcano et al.\ 2024).
In continuing FetchReach\_0.1, the distance threshold for reaching the goal was increased from $0.05$ to $0.1$.
Similarly to continuing Reacher, the position of the goal in continuing FetchReach and continuing PointMaze only changes when the goal is reached, and environment resets only change the position of the agent's embodiment, which are the Fetch robot and the green ball, respectively.
The reward is 100 for the time step that the goal is reached and zero for all other time steps.
To mimic the spatial boundary limitations when learning on real robots, we limit the movement of the Fetch robot's end-effector to be within an imaginary box the same size as and on top of the table.

The \emph{sparse HalfCheetah, InvertedDoublePendulum}, and \emph{InvertedPendulum} tasks were adapted from the corresponding tasks in Gym Mujoco (Brockman et al.\ 2016).
In sparse HalfCheetah, the agent receives a reward of 100 or -100 for every forward or backward movement, respectively, more than three meters.
In sparse InvertedDoublePendulum, the reward is one for every time step that the fingertip has an altitude higher than one.
In sparse InvertedPendulum, the reward is one for every time step that the pole is within 0.2 radians of the vertical z-axis.
Rewards are zero for all other time steps.
The range of motion of the simulated cheetah robot is limited within -1.2 and 1.2 radians so that the robot does not flip over.
Sparse InvertedDoublePendulum and InvertedPendulum do not use state-based resets and only use time-based resets.

As seen in Figure \ref{fig:sparse_tasks}, none of the tested sparse-reward tasks fail to learn without resets.
The performance of the $\tilde{\alpha}$-toggle-LN intervention with $\tilde{\alpha}=0.02$ is comparable to the performance of learning with resets on all tasks except continuing FetchReach and continuing PointMaze.
The $\tilde{\alpha}$-toggle intervention fails to learn in continuing PointMaze and sparse InvertedPendulum.
The layer norm intervention recovers the lost performance when learning without resets in continuing FetchReach, but learns slowly in sparse HalfCheetah.

\clearpage
\subsection{The Visual Ball Pull-Up Task} \label{app:ball_pull_up}

The task consists of a 2D ball and a horizontal boundary line drawn on the monitor in front of the arm.
The goal of the task is to keep the ball above the fixed boundary line for as long as possible.
However, the ball is subject to the force of gravity, which, in the absence of other forces, always pulls the ball down below the boundary.
The ball can be pulled up using a 2D ring, known as the cursor, that can be moved around the screen.
The movement of the cursor in any direction applies a force on the ball in that direction.
To pull the ball above the boundary, the cursor needs to get close to the ball and move up.
The position of the cursor reflects where the robot arm's end-effector is pointing.
Specifically, The position of the cursor on the screen is determined by extending a straight line from the end-eﬀector and perpendicular to its plane and colliding it with the monitor’s plane.
Overall, the arm can move the ball around by pointing its end-eﬀector at the ball and moving it around.
The observations of this task include arm proprioception and images from a camera mounted on the end-eﬀector.
Reward is one when the ball is above the boundary line and -10 otherwise.
The task can be learned with time-based resets every 250 time steps or without.
When learning with time-based resets, the arm is reset to a fixed initial position where the camera and the end-eﬀector are pointing at the monitor.

\end{document}

%% file: math_commands.tex
\usepackage{amsmath,amsfonts,bm}

\def\eqref#1{equation~\ref{#1}}

\def\1{\bm{1}}

\DeclareMathAlphabet{\mathsfit}{\encodingdefault}{\sfdefault}{m}{sl}
\SetMathAlphabet{\mathsfit}{bold}{\encodingdefault}{\sfdefault}{bx}{n}

\DeclareMathOperator{\sign}{sign}